\documentclass[lettersize,journal]{IEEEtran}
\usepackage{amsmath,amsfonts}
\usepackage{algorithmic}
\usepackage{algorithm}
\usepackage{array}
\usepackage[caption=false,font=normalsize,labelfont=sf,textfont=sf]{subfig}
\usepackage{textcomp}
\usepackage{stfloats}
\usepackage{url}
\usepackage{color}
\usepackage{verbatim}
\usepackage{graphicx}
\usepackage{cite}
\usepackage{amssymb}
\usepackage{multirow}
\usepackage{multicol}
\usepackage{booktabs}
\usepackage{framed}
\usepackage{xcolor}
\usepackage{makecell}
\usepackage[normalem]{ulem} 
\newcommand{\eg}{\textit{e}.\textit{g}. }
\newcommand{\etal}{\textit{et al}. }

\newcommand{\mc}[1]{\mathcal{#1}}

\def\eg{\emph{e.g. }}
\def\etal{\emph{et al. }}

\begin{document}

\title{UWStereo: A Large Synthetic Dataset for Underwater Stereo Matching}

\author{Qingxuan Lv, 
       Junyu Dong,~\IEEEmembership{Member,~IEEE,}
      Yuezun Li,~\IEEEmembership{Member,~IEEE,}
      Sheng Chen,~\emph{Life Fellow, IEEE}
      Hui Yu,~\IEEEmembership{Senior Member,~IEEE,}
      Shu Zhang,
      Wenhan Wang

\thanks{{\em Corresponding authors}: Yuezun Li and Junyu Dong} %
\thanks{Q Lv, J Dong, Y Li, S Zhang, and W Wang are with the Department of Computer Science and Technology, Ocean University of China, Qingdao, Shandong Province, 266100 China.}
\thanks{S. Chen is with School of Electronics and Computer Science, University of Southampton, Southampton SO17 1BJ, U.K., and also with the College of Computer Science and Technology, Ocean University of China, Qingdao 266100, China (E-mail: sqc@ecs.soton.ac.uk).} %
\thanks{H. Yu is with School of Psychology \& Neuroscience, University of Glasgow, Glasgow, U.K. (E-mail: Hui.Yu@glasgow.ac.uk).} %
}

\markboth{Journal of \LaTeX\ Class Files,~Vol.~14, No.~8, Octuber~2022}%
{Lv \MakeLowercase{\textit{et al.}}: UWStereo: A Large Synthetic Dataset for Underwater Stereo Matching}

\maketitle

\begin{abstract}
Despite recent advances in stereo matching, the extension to intricate underwater settings remains unexplored, primarily owing to: 1) the reduced visibility, low contrast, and other adverse effects of underwater images; 2) the difficulty in obtaining ground truth data for training deep learning models, i.e. simultaneously capturing an image and estimating its corresponding pixel-wise depth information in underwater environments. To enable further advance in underwater stereo matching, we introduce a large synthetic dataset called \textbf{UWStereo}. Our dataset includes 29,568 synthetic stereo image pairs with dense and accurate disparity annotations for left view. We design four distinct underwater scenes filled with diverse objects such as corals, ships and robots. We also induce additional variations in camera model, lighting, and environmental effects. In comparison with existing underwater datasets, UWStereo is superior in terms of scale, variation, annotation, and photo-realistic image quality. To substantiate the efficacy of the UWStereo dataset, we undertake a comprehensive evaluation compared with nine state-of-the-art algorithms as benchmarks. The results indicate that current models still struggle to generalize to new domains. Hence, we design a new strategy that learns to reconstruct cross domain masked images before stereo matching training and integrate a cross view attention enhancement module that aggregates long-range content information to enhance the generalization ability. 
\end{abstract}


\section{Introduction}\label{sec:intro}
\IEEEPARstart{T}{oward} building an intelligent agent in real world, the ability of visual understanding occupies a significant position in the whole blueprint, especially for parsing 3D scene structure, since it helps the agent locate itself and interact with surroundings \cite{Stoiber_2022_CVPR_3dtrack, lee2022instancewise}. As a primary task for restoring spatial geometry from 2D images, stereo matching have been attracting more and more interests in recent years \cite{chang2018pyramid_psm, lipson2021raftstereo, gangwei2023iterative_igev}, and facilitating the development of many down-stream tasks like Multi-view Stereo \cite{wang2022itermvs, zhang2023geomvsnet}, RGBD-SLAM \cite{wang2023coslam}, 3D reconstruction \cite{ren2023volrecon}, and etc.

\begin{figure}[t]
  \centering
  \includegraphics[width=\linewidth]{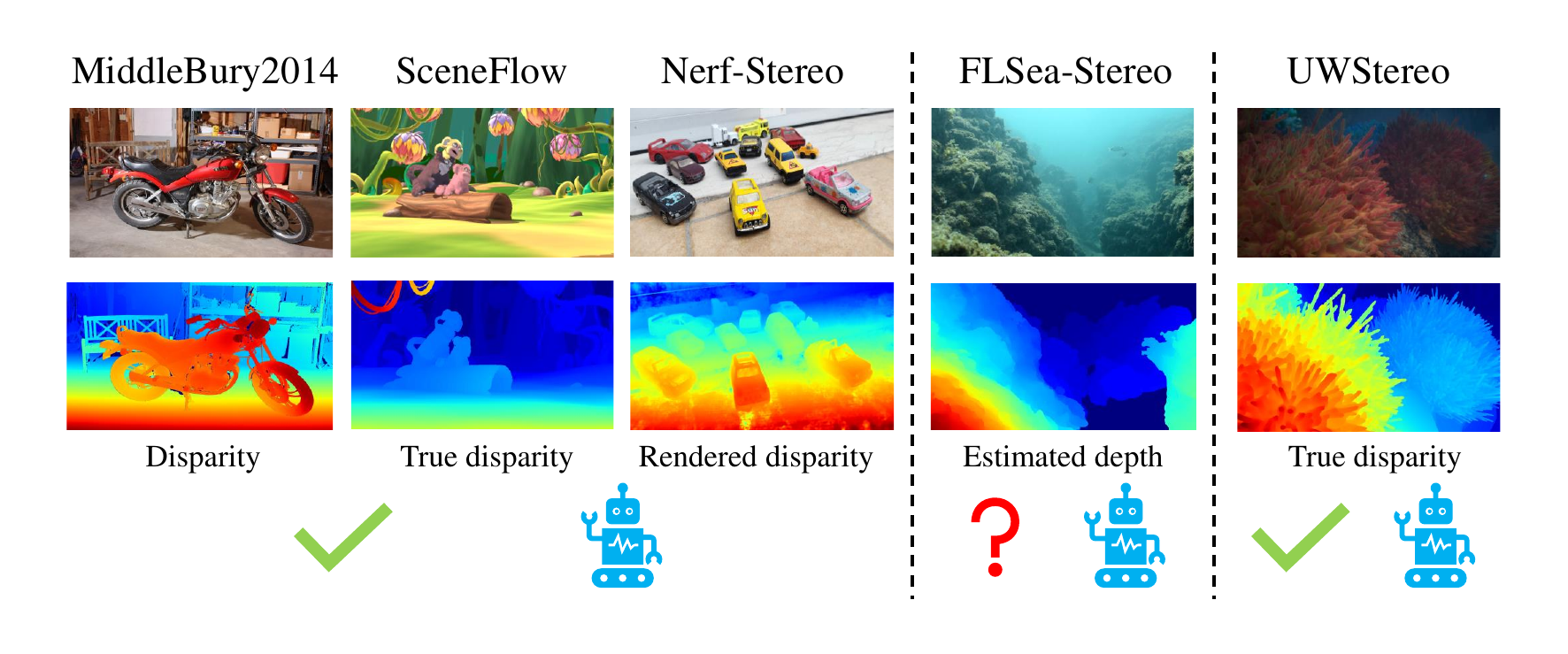}
  \caption{Illustration of the dilemma for underwater stereo matching. \textit{Left}: With sufficient datasets, stereo matching models can be easily trained, evaluated, and applied on aquatic environments. \textit{Middle}: The accurate depth information is hard to acquired in real underwater scenes. \textit{Right}: Our UWStereo is able to provide accurate depth information for all pixels and synthesize photo-realistic underwater images.}
  \label{fig:intro}
\end{figure}

Existing methods show two main branches for the development of stereo matching models: one focuses on direct disparity prediction through cost aggregation extension, exemplified by works such as \cite{chang2018pyramid_psm, xu2022attention, xu2020aanet, guo2019groupwise_gwc}, while the other employs iterative optimization to refine disparity estimates, demonstrated in contributions like \cite{lipson2021raftstereo, gangwei2023iterative_igev, zhao2023highfrequency_hsm, li2022practical_cres}. Most of them adhere to the simulation-to-real (Sim2Real) training paradigm to avoid the data scarcity. This paradigm involves initially training the model on extensive synthetic datasets, \eg SceneFlow \cite{mayer2016a_sceneflow} and Nerf-Stereo \cite{tosi2023nerfsupervised}, for depth recovery, followed by fine-tuning on real-world datasets like KITTI \cite{kitti2012, kitti2015}, MiddleBury2014 \cite{Middlebury}, and ETH 3D \cite{schops2017multi_eth3d}. This training approach has been proven to be effective (left part of Fig.~\ref{fig:intro}), yet its application in underwater settings, as discussed in \cite{himbdataset, berman2020underwater_squid, randall2023flsea, uieb}, faces considerable challenges.

\begin{table*}[t]
  \centering
\caption{Statistics of existing stereo matching datasets.}
  \label{tab:datasetstat}
  \small
  \begin{tabular}{lccccccccccc}
    \toprule
    Dataset & \makecell{Training \\ Images }& \makecell{Testing \\ Images} & Resolution & \makecell{Stereo \\ Disparity} & Underwater & Baseline & \makecell{Focal \\ Length}\\
    \midrule
    MPI Sintel \cite{mpisintel} & 1064 & 564 & 1024$\times$436 & Dense & No & $10 (unit)$ & -\\
    KITTI-2012 \cite{kitti2012} & 194 & 195 & 1226$\times$370 & Sparse & No & - & -\\
    KITTI-2015 \cite{kitti2015} & 200 & 200 & 1242$\times$375 & Sparse & No & - & -\\
    MiddleBury2014 \cite{Middlebury} & 23 & 10 & 2880$\times$1988 & Dense & No & - & -\\
    SceneFLow \cite{mayer2016a_sceneflow} & 35454 & 4370 & 960$\times$540 & Dense & No & $1 (unit)$ & $1050,450$ \\
    Nerf-Stereo \cite{tosi2023nerfsupervised} & 65148 & - & 1160$\times$522 & Rendered & No & \makecell{$0.5, 0.3,$ \\$ 0.1 (cm)$} & -\\
    HIMB \cite{himbdataset} & 4047 & 15 & - & No & Yes & - & -\\
    FLSea-Stereo \cite{randall2023flsea} & 7337 & - & 1280$\times$720 & \makecell{Estimated \\ Depth} & Yes & - & $\approx1800$\\
    VAROS \cite{zwilgmeyer2021the_varos} & 4713 & - &  1280$\times$720 & No & Yes & - & -\\
    UWStereo & 26612 & 2956 & 1280$\times$720 & Dense & Yes & \makecell{$6, 12, 18,$\\$ 24, 30 (unit)$} & $1400$ \\
    \bottomrule
  \end{tabular}
\end{table*}

Basically, the quality of images captured in underwater scenes is significantly degraded due to factors like scattering, light absorption, and refraction \cite{akkaynak2018a}. These issues result in reduced visibility, low contrast, and various adverse effects, thereby limiting the practical utility of underwater data in oceanic engineering applications. Additionally, the lack of effective sensors and solutions for accurately estimating depth information in underwater environments makes it impossible to obtain images with corresponding ground truth disparity annotations. These formidable challenges underscore the current predicament in underwater stereo matching research \cite{tcsvt_sm_uw_3}. As illustrated in the middle portion of Fig.~\ref{fig:intro}, the estimated depth map, serving as the ground truth in the FLSea-Stereo dataset \cite{randall2023flsea}, exhibits compromised accuracy when compared to analogous data in existing datasets.

\begin{figure*}
    \centering
    \includegraphics[width=0.9\linewidth]{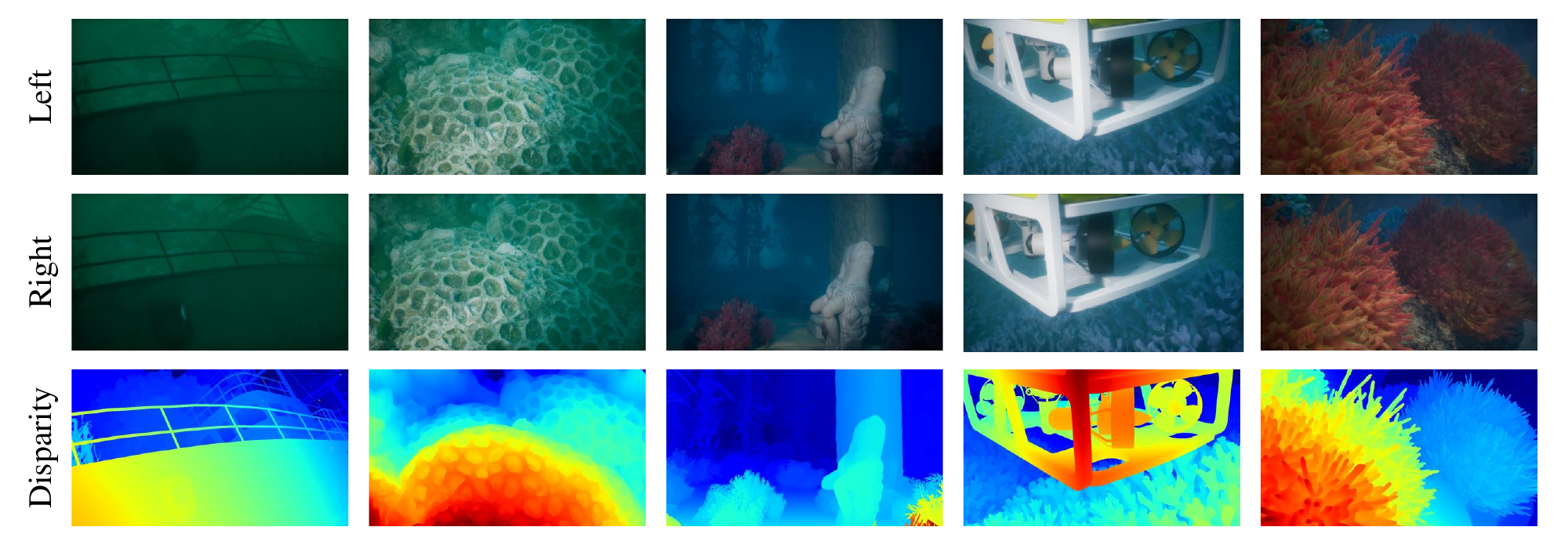}
    \caption{Stereo image examples from the UWStereo dataset.}
    \label{fig:example}
\end{figure*}

To address the challenges of scarcity of underwater stereo matching data, we present a novel dataset named \textbf{UWStereo} in this paper. To simulate intricate underwater environments, we utilize Unreal Engine 5 (UE5\footnote{UE5: \url{https://www.unrealengine.com/}}) to create four distinct synthetic scenes, namely coral, default, industry, and ship, and fill them with diverse objects including corals, rocks, ships, and etc. In detail, the UWStereo comprises 29,568 stereo image pairs with accurate disparity annotations. Like MPI Sintel \cite{mpisintel} and Nerf-Stereo \cite{tosi2023nerfsupervised}, we also incorporate extra variations in camera baseline, light sources, and other environmental effects. Compared to existing datasets, UWStereo is the first synthetic dataset toward underwater scenes and has several advantages: 1) it has a large number of images and low redundancy; 2) it contains rich samples at a close view to underwater objects for underwater scenes; 3) the pixel-level annotations are dense and accurate. A detailed comparison between UWStereo and other existing datasets is presented in Table~\ref{tab:datasetstat} and we show some examples in Fig.~\ref{fig:example}.

Leveraging this dataset, we conduct an extensive benchmark by selecting nine recently developed stereo matching algorithms and retraining them. The results underscore the superiority of iterative optimization methods over other approaches. We further evaluate the generalization ability of these models. The comparison reveals that existing stereo matching methods are still hard to generalize to underwater scenes. To overcome this, we first refine an iterative stereo matching model by integrating a cross view attention enhancement module to enable the model to aggregate long-range content information from stereo images. Then, we draw inspiration from masked image learning and design a new training strategy by employing a paired masked image reconstruction pretraining task before stereo matching training. The results indicate that recovering paired masked images with a small mask ratio will enhance the generalization ability. We hope our work can inspire further research interests for underwater stereo matching and other down-stream tasks. The contributions of this paper are summarized as:

\begin{enumerate}
    \item We introduce a large synthetic stereo matching dataset containing 29,568 stereo images with dense and accurate disparity annotations, aiming to facilitate the researches toward developing stereo matching networks for real-world underwater scenes. 
    \item We select and retrain nine recent stereo matching algorithms for the benchmarking purpose and perform evaluation for their generalization ability.
    \item To enhance the generalization ability between terrestrial and underwater synthetic datasets, We induce a cross view enhancement module that enable the model to aggregate long-range context information, and design a new strategy that couples the stereo matching with a paired masked image reconstruction pretraining stage. 
\end{enumerate}

The remainder of this paper is organized as follows. In Section II, some related works is briefly introduced. Section III presents the details of synthesizing the underwater stereo matching dataset. Section VI demonstrates the training strategy for developing generalized stereo matching network. Section V contains the experimental results and discussions about the value of the proposed dataset. Finally, Section VI concludes this paper.

\section{Related Works}
\label{sec:relatedwork}
\subsection{Stereo Matching}
Traditional stereo matching algorithms typically consist of four key steps including matching cost computation, cost aggregation, optimization, and disparity refinement \cite{scharstein2002taxonomy}. With the success of deep neural networks, early attempts focus on how to compute the accurate matching cost \cite{zbontar2016stereo, shaked2017improved}. Further studies turn to consider the post-processing of the disparity estimation \cite{guney2015displets, seki2017sgm_net}. Recently, many works emerged with an efficient end-to-end structure. One prominent way is to employ 3D convolutional networks to regularize and filter the cost volume to improve the representative ability of the cost volume \cite{chang2018pyramid_psm, guo2019groupwise_gwc, xu2022attention, xu2020aanet, zhang2020domaininvariant_dsm, tcsvt_sm_1, tcsvt_sm_2}. Differently, inspired by the optical flow estimation \cite{teed2020raft}, another branch introduced a novel iterative optimization strategy to iteratively optimize the disparity field using recurrent convolutional structure \cite{lipson2021raftstereo, gangwei2023iterative_igev, li2022practical_cres, zhao2023highfrequency_hsm}. Specifically, RAFT-Stereo \cite{lipson2021raftstereo} firstly introduced the multi-level GRU units to update the disparity map. Then, IGEV \cite{gangwei2023iterative_igev} extended RAFT-Stereo with a combined geometry encoding volume, which provides concise initial disparity for updating instead starting with a blank map. HSMNet \cite{zhao2023highfrequency_hsm} proposed a decouple LSTM module to keep high-frequency information during the iterative updating stage. CREStereo \cite{li2022practical_cres} proposed a cascaded recurrent network to independently refine disparities in different cascade level.

Notably, some works try to incorporate self-supervised learning into the framework of stereo matching \cite{tosi2023nerfsupervised, rao2023masked_mask, Weinzaepfel_2023_ICCV_croco}. Nerf-Stereo \cite{tosi2023nerfsupervised} rendered a large amount of image triplets by leveraging neural radiance field (Nerf) \cite{mildenhall2020nerf} and developed triplet photometric loss and rendered disparity loss to train the model with self-supervised signals. Rao \etal \cite{rao2023masked_mask} formed a multi-task learning paradigm by simultaneously training the model with stereo matching and masked image reconstruction. Croco-Stereo \cite{Weinzaepfel_2023_ICCV_croco} performed self-supervised cross-view completion pretraining to encourage the model to learn dense geometric information.

\subsection{Stereo Matching Datasets}

In the field of stereo matching, several widely used datasets, such as Middlebury2014 \cite{Middlebury}, KITTI \cite{kitti2012, kitti2015}, and ETH3D \cite{schops2017multi_eth3d}, have become important benchmarks in research. These datasets have advanced the development and optimization of algorithms by providing high-precision depth information. However, a common limitation of these datasets is their relatively small size, primarily due to the challenges of obtaining pixel-wise depth information. In real-world scenarios, manually annotating depth information is not only time-consuming and labor-intensive but also prone to environmental variations such as lighting, reflections, and occlusions, which further complicate and increase the cost of annotation.

To address these challenges, researchers have begun exploring the emerging simulation-to-real paradigm. This approach leverages rendered images and depth information in synthetic scenes to overcome the limitations of scarce real-world data. MPI Sintel \cite{mpisintel} is one of the pioneers in this field, synthesizing over 1,000 stereo images with depth maps from open-source animation clips. Following this, the SceneFlow \cite{mayer2016a_sceneflow} dataset marked a significant advancement in the field by introducing a synthetic dataset containing more than 35,000 stereo frames, involving over 30,000 objects extracted from ShapeNet \cite{savva2015semantically}. The diverse appearances of these objects greatly enrich the variety of training data, effectively addressing the challenge of training models with limited real-world data.

These synthetic datasets not only surpass traditional datasets in terms of data volume but also provide highly controllable environments, allowing researchers to systematically explore the performance of various algorithms. However, despite the clear advantages of synthetic datasets in enriching data diversity, the domain gap between synthetic and real-world scenarios remains a critical issue. Consequently, many studies are now focusing on how to transfer the knowledge learned from synthetic data to real-world scenes, further narrowing the gap between simulation and reality. These efforts not only help enhance the robustness of models in real-world environments but also lay the foundation for the future development of more broadly applicable stereo matching systems.

\subsection{Underwater Image Datasets}

Several datasets have been developed to support various vision tasks within the underwater domain. For example, the UIEB \cite{uieb} dataset was introduced specifically for underwater image enhancement, providing a benchmark for improving the visual quality of underwater imagery. The SUIM \cite{islam2020semantic_suim} dataset goes further by offering underwater images with pixel-level semantic annotations, making it useful for tasks such as object detection and segmentation in underwater environments. Additionally, the SQUID \cite{berman2020underwater_squid} and Sea-thru \cite{akkaynak2019seathru} datasets provide raw underwater images along with estimated distance maps, which are crucial for depth estimation and 3D reconstruction tasks.

Building on these efforts, the FLSea dataset \cite{randall2023flsea} advances the field by offering both monocular and binocular images paired with estimated depth information, making it a valuable resource for stereo matching and depth prediction tasks. The VAROS dataset \cite{zwilgmeyer2021the_varos}, another significant contribution, extends the synthetic pipeline for underwater environments by providing 4,713 synthetic monocular images with corresponding depth maps and surface normal information. These datasets collectively highlight the importance of recovering depth information from underwater images for various vision tasks, including image restoration, reconstruction, and 3D modeling.

Despite these advancements, existing underwater datasets still face several limitations. Many datasets, such as UIEB, SQUID, and HIMB, are relatively small in scale, which limits the diversity and complexity of the scenes they cover. Datasets like VAROS and FLSea, while larger, often feature simpler scene structures, which may not fully capture the complexity of real-world underwater environments. Additionally, some datasets, including SQUID, FLSea, and Sea-thru Nerf \cite{akkaynak2019seathru}, suffer from compromised or less accurate distance annotations, which can impact the performance of depth-related tasks.

To address these challenges, we propose a new synthetic stereo matching dataset specifically designed for underwater scenes. Our dataset is characterized by its substantial size, diverse scene variations, and highly accurate annotations, setting it apart from previous datasets. By offering a more comprehensive and reliable resource, our dataset aims to stimulate further research in the field of underwater stereo matching, ultimately contributing to more robust and effective underwater vision systems.



\section{The UWStereo Dataset}\label{S3}
\subsection{Synthesizing Workflow}
\begin{figure}
    \centering
    \includegraphics[width=1\linewidth]{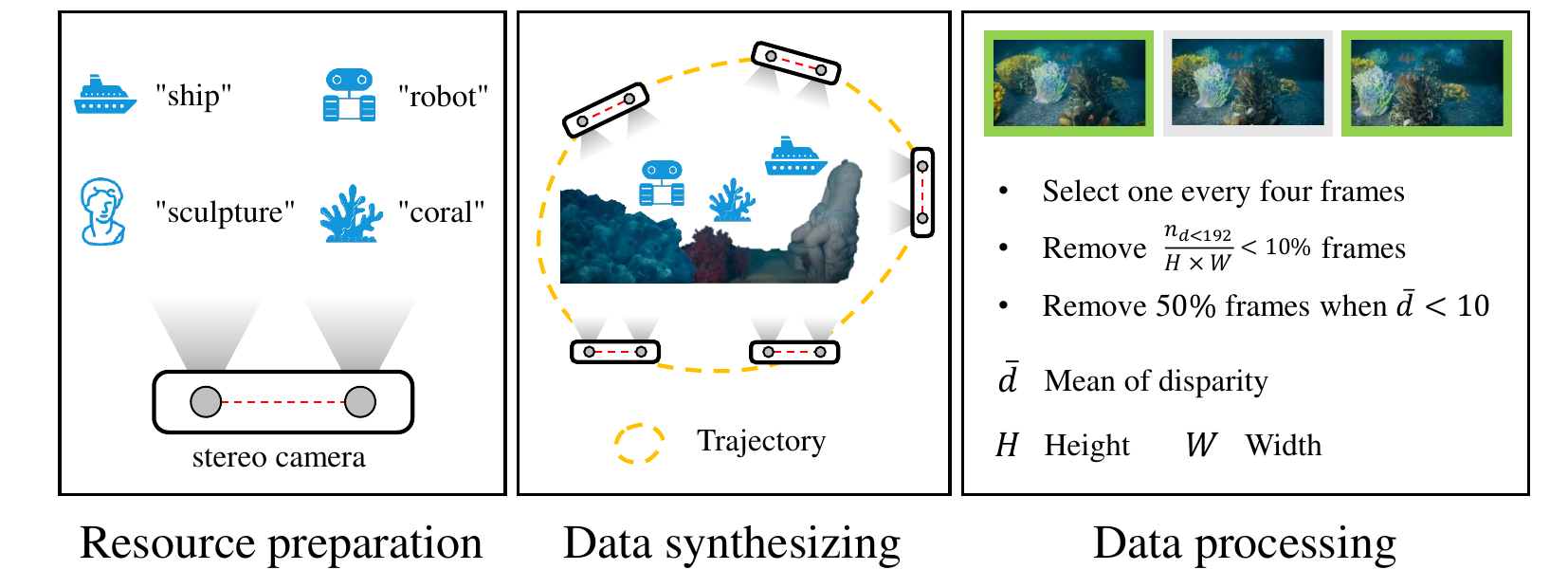}
    \caption{Synthesizing workflow.}
    \label{fig:workflow}
\end{figure}

Previous works synthesized underwater images by using Blender\footnote{Blender: \url{https://www.blender.org/}} rendering engine \cite{mayer2016a_sceneflow, zwilgmeyer2021the_varos} or generative models like Nerf \cite{tosi2023nerfsupervised} and Generative Adversarial Network (GAN) \cite{li2017watergan}. However, it's not trivial to capture underwater stereo images with accurate dense depth information for above methods, as the insufficient ability of generative models in rendering depth information and the distribution discrepancy between synthetic and real-world images. Thus, toward providing dense and accurate depth information for underwater scenes, we employ UE5 as the simulator because of its powerful ability in simulating real-world visual effects \footnote{We use an open source plugin named EasySynth \url{https://github.com/ydrive/EasySynth} for data generation}. The synthesizing workflow contains three steps and is shown in Fig.~\ref{fig:workflow}.

\noindent\textbf{Resource preparation:}
MPI Sintel \cite{mpisintel} utilizes open-source animation sequences as simulation environments, while SceneFlow \cite{mayer2016a_sceneflow} uses digital objects provided by some animations or ShapeNet \cite{savva2015semantically}. These virtual assets empower the synthetic data to provide sufficient variation of disparity distributions. Inspired by these works, we invested substantial efforts in collecting numerous virtual resources that could potentially appear in underwater scenes to enrich the disparity distributions. The collected resources includes \textit{sunken shipwrecks, scanned coral, simple sculpture models, dilapidated cars, wind turbine generators, and underwater robots}. Based on these diverse objects, we constructed four distinct underwater scenes named coral, default, industry, and ship respectively, denoting as $S_i$.

we also design another crucial component, a binocular camera, simulated by using two separate camera components. Similar to Sceneflow, we set the film back size to $32mm\times18mm$ and focus plane $f_p = 150$ units in simulator. The two camera components have the same intrinsic parameters, denoting as:
\begin{equation}
\small
    \textbf{K} = \left[ \begin{array}{ccc}
        f_x &  0 & c_x \\
        0 & f_y  & c_y\\
        0 &   0  & 0
    \end{array} \right]  = \left[ \begin{array}{ccc}
        1400 &  0 & 640 \\
        0 & 1400  & 360\\
        0 &   0  & 0
    \end{array} \right].
\end{equation}
We only consider translation between two camera, where the translation vector is $\textbf{t} = [b,0,0]$. Differently, we regard the baseline $b$ as a variable rather than a fixed value, which is also examined in DSMNet \cite{zhang2020domaininvariant_dsm} or Nerf-Stereo \cite{tosi2023nerfsupervised}.

\noindent\textbf{Data synthesizing:}
Having prepared the requisite virtual assets, we proceed to the data synthesizing phase. In the context of each $S_i$, an initial camera position $p$ is selected in proximity to visible objects, aiming to facilitate the subsequent camera motion. We then create an animation sequence along with an manually edited motion trajectory that emulates the movement of a binocular camera. Notably, we observed that the sequence lengths in both MPI Sintel (up to $50$) and SceneFlow (up to $300$) datasets are relatively limited. In contrast, we set the frame rate to 15 frame per second and edit a long enough sequence with at least $1500$ frames for each scene. The ground truth disparity is generated according to:
\begin{equation}
\small
    disparity = \frac{b\cdot f_x}{depth},
\end{equation}
where $b$ and $f_x$ can be retrieved from the camera parameters, and depth information is synthesized during moving. 

Furthermore, to introduce additional variations, we exercise control over synthesized data by manipulating key factors such as camera baseline, light sources, and volumetric fog. Specifically, under the assumption of camera sensor has a restricted visual range in underwater scenes, we assign the baseline $b$ to a certain value within the set $\{6, 12, 18, 24, 30\}$ to generate balance disparity distribution. The utilization of additional light sources, represented as $E_l \in \{0, 1\}$, is also considered since binocular camera sensors are commonly integrated into remotely operated vehicles (ROVs) equipped with additional lighting sources. Moreover, volumetric fog, a significant feature of UE5, is harnessed to simulate underwater light scattering conditions. Within this context, we diversify the visual attributes of the synthesized data by manipulating the density and color of the volumetric fog. The density of volumetric fog is specified as $E_d \in \{1.0, 2.0\}$, and the color variation includes blue and green, denoted as $E_c \in \{blue, green\}$. These diverse settings yield the synthesis of approximately $60,000$ consecutive stereo matching data instances derived from a single trajectory within a scene, calculated as $1500 \times 5 \times 2 \times 2 \times 2 = 60,000$. Given that we have incorporated four distinct scenes, the cumulative volume of synthesized data characterized by precise depth information approximates to around $240,000$ instances.

\begin{table}[t]
\scriptsize
  \centering
  \vspace{-0.2cm}
  \caption{Details of the UWStereo.}
  \vspace{-0.2cm}
  \label{tab:datadetail}
  \scriptsize
  \begin{tabular}{lcccc}
    \toprule
    Dataset  & Total Images & Training Images & Testing Images \\
    \midrule
    coral     & 4,494 & 4,045 & 449  \\
    default   & 4,720 & 4,248 & 472  \\
    industry  & 9,230 & 8,307 & 923  \\
    ship      & 11,124 & 10,012 & 1,112 \\
    all       & 29,568 & 26,612 & 2,956\\
    \bottomrule
  \end{tabular}
\end{table}

\noindent\textbf{Data processing:}
Previous datasets employed continuous frames to guarantee the continuity. Despite the large size, those synthetic data are redundant in the temporal dimension leading to longer training costs. Another important problem is the synthetic disparity distribution may be imbalanced, as the camera motion trajectory need to be manually edited, which inevitably induces motion bias. 

Considering to resolve the above perturbations, we decide to design a further data processing step to cleanup the synthetic data. Concretely, we design three rules as: 1. Set interval to 4 to sample a frame. 2. Remove 50\% of the frames which satisfy $\bar{d} < 10$, where $\bar{d}$ represents the mean of disparities. 3. Remove the data when $\frac{n_{disp>192}}{H\times W} > 10\% $, where $n_{disp>192}$ represents the number of pixels that the disparity exceeds 192. $H$ and $W$ denote the height and width of the image. These rules help remove excessive imbalanced data, facilitating the convergence of training (the comparison of disparity distributions between our UWStereo and other datasets is presented in supplementary material.).

\subsection{Dataset Details}
The proposed UWStereo dataset contains a total of 29,568 stereo matching image pairs oriented to the underwater environment. All data contain dense disparity annotations for left view. The details of the dataset are presented in Table~\ref{tab:datadetail}. We randomly selected 10\% of each scene to serve as testing images and the remaining 90\% as training data. In comparison with existing stereo matching datasets, UWStereo is firstly designed for intricate underwater scenes and contains several advantages: 1) it has large number of images and low redundancy; 2) it contains rich samples at close view to underwater objects instead the distant views for underwater scenes; 3) the pixel-level annotations are dense and accurate.

\begin{figure*}
    \centering
    \includegraphics[width=1.0\linewidth]{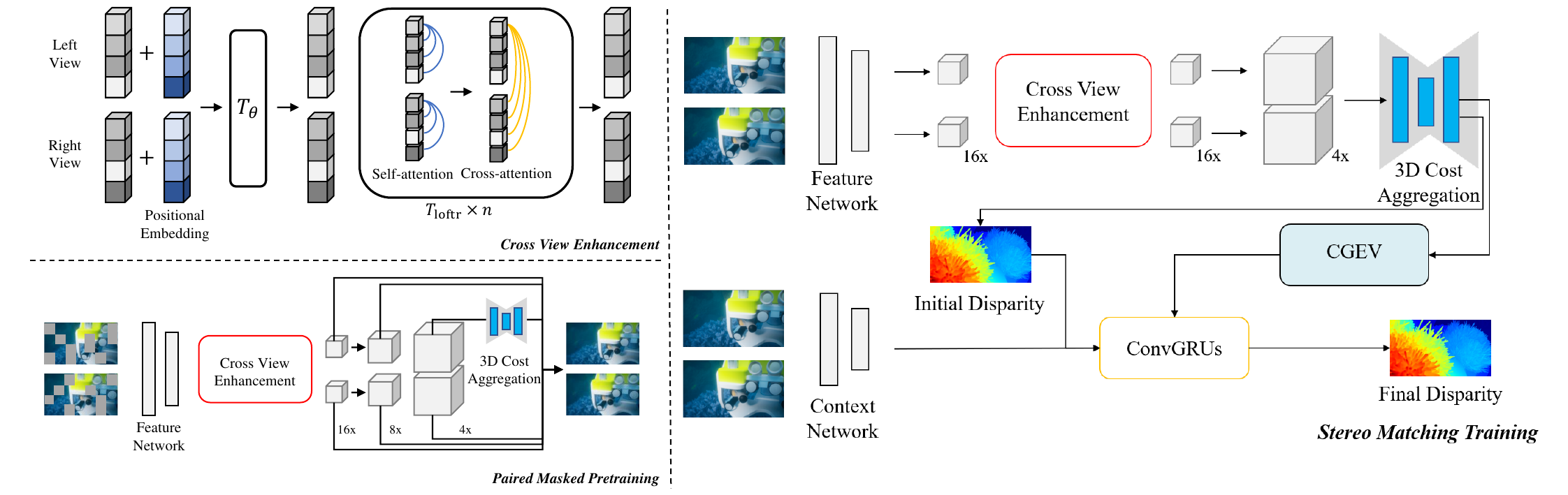}
    \caption{\textit{Left top}: The structure of Cross View Enhancement (CVE) module. \textit{Left bottom}: The network structure employed during pretraining. \textit{Right}: The network structure for stereo matching training.}
    \label{fig:modelmask}
    \vspace{-0.6cm}
\end{figure*}

\section{The Strategy}\label{S4}
We notice that some recent works \cite{Weinzaepfel_2023_ICCV_croco, rao2023masked_mask, zhu2023pmatch} effectively relate the masked representation learning \cite{Doll2022Masked_mae} to the task of cross-image matching, which suggests that pretraining the model with masked image reconstruction objectives improves the generalization ability. Inspired by these works, we aim to build up a paired masked image reconstruction pretext task on both terrestrial and underwater images to study whether the masked representation can enhance the generalization ability.

\noindent
\textbf{Network Structure}
We build our model based the IGEV \cite{gangwei2023iterative_igev}. One difference is that we insert a new cross view enhancement (CVE) module behind the feature extractor, since we observe the correspondence between two different views helps the model to perceive geometry structure of scene \cite{zhu2023pmatch, sun2021loftr}. Concretely, the CVE module consists of a Multi-Head Attention block (MHA) and $n$ LoFTR blocks denoting as $T_\theta$ and $T_{loftr}$ respectively. $T_\theta$ is used to adjust the CNN-extracted representation to patch-wise style. Following PMatch \cite{zhu2023pmatch}, we also add positional embeddings before feeding to $T_{loftr}$ to encourage the model to track the positional information for each patch. Then, since the LoFTR block is able to aggregate long-range context information by self and cross attention layers, we employ multiple $T_{loftr}$ layers to focus on searching cross view correspondence to enhance the feature representation. The left top part of Fig.\ref{fig:modelmask} illustrate the structure of CVE. The network structures for pretraining and stereo matching training are also presented in the left bottom and right parts of Fig.~\ref{fig:modelmask}. The cost aggregation module is kept during pretraining, as the matching cost can also provide guidance to recover paired images. While, we drop the iterative updating module in pretraining stage, as it is not suit for paired masked image reconstruction pretext task. We design some lightweight decoders at different scales to reconstruct the masked images.

\noindent
\textbf{Pretraining setting}
Unlike Masked-CFNet \cite{rao2023masked_mask} which introduces a multi-task learning framework, we follow PMatch \cite{zhu2023pmatch} to form a self-supervised pretext task by reconstructing the paired images. Given left and right images denoting as $x_l$ and $x_r$, we feed them to feature extractor to extract features at scale $s = 2$ denoting as $f_l^{s=2}$ and $f_r^{s=2}$. We then use two predefined mask ratio $r_1$ and $r_2$ to randomly generate masks, where the patch size is $32\times32$ instead a small mask size in Masked-CFNet. The masked feature can be represented as:
\begin{equation}
    f_l^{s=2'} = f_l^{s=2} * (1 - w) + m * w,
\end{equation}
where $m$ represents the mask tokens and $w$ is the corresponding mask.
The masked features are then fed to the following modules to extract features at $s=4$, $s=8$, and $s=16$. Furthermore, we reconstruct the input images by regressing the raw pixel values for feature at each scale. It should be noticed that the cost aggregation is appended for left feature at $s=4$, since we want to keep it to be consistent with the stereo matching training. The reconstruction objective can be formed as:
\begin{equation}
    \mc{L}_M = \sum_s\frac{1}{N_{mask}}(||x_l^s - x_l^{s'}||_1 + ||x_r^s - x_r^{s'}||_1),
\end{equation}
where $N_{mask}$ denotes the number of masked image patches. $x_l^{s'}$ and $x_r^{s'}$ denote the reconstructed images at a scale $s$.

\noindent
\textbf{Stereo Matching Training}
Following previous works \cite{chang2018pyramid_psm, lipson2021raftstereo, gangwei2023iterative_igev, zhao2023highfrequency_hsm}, we employ $l_1$ loss as the objective of stereo matching. Like IGEV \cite{gangwei2023iterative_igev}, we supervise the network on both initial disparity and updated disparity map. Hence, the training loss can be defined as:
\begin{equation}
\begin{aligned}
    \mc{L}_T &= SmoothL1(d_{init}-d_{gt}) \\
    &+ \sum_{i=1}^{N_{iter}} \gamma^{N_{iter} - i}||d_i - d_{gt}||_1,
\end{aligned}
\end{equation}
where $\gamma = 0.9$ and $N_{iter}$ denotes the number of iteration. 



\begin{table*}[t]
  \centering
  \small
  \caption{Comparison with state-of-the-art methods.}
\label{tab:benchmark}
\begin{tabular}{lcccccccccc}
\toprule
\multirow{2}{*}{Method}  & \multicolumn{2}{c}{Coral} & \multicolumn{2}{c}{Default} & \multicolumn{2}{c}{Industry} & \multicolumn{2}{c}{Ship} & \multicolumn{2}{c}{All} \\
     & EPE          & $>$3px    & EPE          & $>$3px    & EPE          & $>$3px   & EPE          & $>$3px    & EPE          & $>$3px    \\
\midrule
LEAStereo \cite{cheng2020hierarchical_lea}  & 2.32 & 12.02 & 0.87 & 4.35 & 1.83  & 10.02  & 1.94 & 8.12 & 1.49 &7.06\\
PSMNet \cite{chang2018pyramid_psm}      & 2.65  & 13.88 & 1.26  & 5.96  & 1.68  &8.42  & 1.62 &  7.50 & 1.32 & 6.75\\
AANet \cite{zhang2019ganet}    & 2.61   &11.92    & 0.97        & 4.04  & 1.31  & 6.65 & 1.52 & 7.20  & 1.27 & 5.81  \\
GwcNet \cite{guo2019groupwise_gwc}    & 2.25      &   10.27 & 0.71   &  2.88 & 1.05   & 4.43 & 1.06  &  4.27 & 1.07 & 4.23  \\
ACVNet \cite{xu2022attention}   & 1.93      & 7.90   & 0.50   &  1.63   & 0.81     &3.28  & 0.79  & 2.95  & 0.85     &  \textbf{3.13}  \\
RAFT-Stereo \cite{lipson2021raftstereo}& 1.55   & 7.86 & 0.43   &   1.57    &0.66    & 3.22   & 0.67    &  3.24   & 0.93   & 4.64    \\

CREStrereo \cite{li2022practical_cres} & 2.07 & 12.47  & 0.37  & 1.34   & -            & -     & \textbf{0.57}  & \textbf{2.55}  & -            & -     \\
HSMNet \cite{zhao2023highfrequency_hsm}   & \textbf{1.09}   & \textbf{6.26}  & \textbf{0.36}  & \textbf{1.29} & \textbf{0.44}  &\textbf{1.94} & 0.59  &2.79 & \textbf{0.68}  &3.25\\
\hline
IGEV \cite{gangwei2023iterative_igev}      & \textbf{1.49}  & \textbf{7.53}  & 0.50  & 1.51  & 0.51 & 2.12 & 0.68 &3.23 & 0.73 & 3.45 \\
Ours        & 1.53 & 7.91  &  \textbf{0.39} & \textbf{1.45} & \textbf{ 0.48}   & \textbf{2.06}  & \textbf{0.62 }  & \textbf{3.01} & \textbf{0.69} & \textbf{3.44 }\\     
\bottomrule
\end{tabular}
\end{table*}

\begin{table*}[t]
  \centering
  \small
\caption{Generalization ability comparison. 'SF' refers to the SceneFlow dataset. '*' means that a mixture of existing datasets are used.}
\label{tab:generalizatinobenchmark}
\begin{tabular}{lccccccccccc}
\toprule
\multirow{2}{*}{Method} & \multirow{2}{*}{Training Set}  & \multicolumn{2}{c}{Coral} & \multicolumn{2}{c}{Default} & \multicolumn{2}{c}{Industry} & \multicolumn{2}{c}{Ship} & \multicolumn{2}{c}{All} \\
    & & EPE          & $>$3px    & EPE          & $>$3px    & EPE          & $>$3px    & EPE          & $>$3px    & EPE          & $>$3px    \\
\midrule
LEAStereo \cite{cheng2020hierarchical_lea} & SF & 2.50 & 12.99 & 1.42 & 5.81 & 1.88 & 9.22  & 3.56 & 14.52 & 2.53 &11.26\\
PSMNet \cite{chang2018pyramid_psm}  & SF  & 2.68  & 13.17 & 2.36  & 7.21 & 2.85  & 10.58  & 4.95 & 15.70 & 3.54 & 12.36\\
AANet \cite{xu2020aanet}  & SF   & 2.94  &12.74  & 1.42   & 7.70  & 1.88  & 12.82 & 3.56 & 19.25  & 2.53 & 14.34  \\
GwcNet \cite{guo2019groupwise_gwc}  & SF  & 2.26 & 11.50 & 1.62 & 6.70 & 1.92 &9.41&4.49 &16.52 &2.89 &11.97 \\
ACVNet \cite{xu2022attention}  & SF  & 2.88 & 13.70 & 3.40 & 8.24 & 3.46  & 11.41 & 7.45  &18.48  & 4.86 & 13.91 \\
RAFT-Stereo \cite{lipson2021raftstereo}& SF & 2.23  & 11.60 & 1.12 & 6.25 & 2.23  & 8.96 & 2.88 & 13.10  & 2.30  & 10.49  \\
CREStrereo \cite{li2022practical_cres} & * & 2.10  & \textbf{9.75}  & 1.19   & 6.05  & \textbf{1.19}        & 6.25  & 3.48  & 12.39  & 2.19  & \textbf{9.06}   \\
HSMNet \cite{zhao2023highfrequency_hsm} & SF  & 2.61  & 13.97  & 5.38  & 10.47 & 4.18  &12.33 & 10.61  &21.52 & 6.55  &15.74\\
IGEV \cite{gangwei2023iterative_igev}   & SF   & 2.29 & 12.60  & \textbf{1.09} & \textbf{5.56}  & 1.99 & 7.58 & \textbf{2.65} & \textbf{12.28} & \textbf{2.14} & 9.74 \\
Ours      & SF   &  \textbf{2.08}  & 10.81   & 1.40  & 5.62  & 1.48 & \textbf{6.13} & 3.86  & 13.10  & 2.45& 9.38 \\ 
\hline
Graft-PSMNet \cite{liu2022graftnet} & SF & 3.27  & 15.31  & 2.87 & 3.23 & 3.91 & 10.48 & 6.37  & 16.13 & 4.57 & 12.82 \\
Croco-Stereo \cite{Weinzaepfel_2023_ICCV_croco}  & * & 2.43  & 12.52 & 2.19 & 9.40 & 2.03 & 9.40 & 4.51 & 16.12 & 3.05 & 12.21 \\
Nerf-PSMNet \cite{tosi2023nerfsupervised} & Nerf-rendered  & 2.48  & 11.87 & 2.26 & 7.82 & 2.07 & 9.47  & 4.63 & 15.82 & 3.13 & 11.96 \\
Nerf-RAFT \cite{tosi2023nerfsupervised}  & Nerf-rendered  & 2.29  & 10.76  & 1.41 & 5.74 & 1.95 & 8.12 & 3.46 &  12.49 & 2.48 & 9.78 \\
DSMNet \cite{zhang2020domaininvariant_dsm} & SF+Carla\cite{pmlr-v78-dosovitskiy17a_carla}  & 2.19  & 11.42  &  1.00   & 4.56  & 1.39  & 6.89 & 2.36 & 10.78  &1.81   & 8.67 \\
\bottomrule
\end{tabular}
\end{table*}

\section{Experiments}\label{S5}

\subsection{Experimental Setting}

\noindent\textbf{Compared Methods}
For benchmarking purposes, we have selected nine stereo matching methods, which are PSMNet \cite{chang2018pyramid_psm}, GwcNet \cite{guo2019groupwise_gwc}, ACVNet \cite{xu2022attention}, AANet \cite{xu2020aanet}, IGEV \cite{gangwei2023iterative_igev}, RAFT-Stereo \cite{liu2022graftnet}, CREStereo \cite{li2022practical_cres}, HSMNet \cite{zhao2023highfrequency_hsm}, LEAStereo \cite{cheng2020hierarchical_lea}. Additionally, we have included two methods specifically designed for generalized stereo matching models, namely GraftNet \cite{liu2022graftnet} and DSMNet \cite{zhang2020domaininvariant_dsm}, in our comparative analysis. Nerf-Stereo \cite{tosi2023nerfsupervised} and Croco-Stereo \cite{Weinzaepfel_2023_ICCV_croco} were also taken into account as their excellent generalization capabilities.

\noindent\textbf{Implementation details}
For our model, we use $4$ LoFTR layer to construct the CVE module. In pretraining stage, we set $r1=r2=0.5$ to generate masks. The training set consists of KITTI \cite{kitti2012, kitti2015}, Middlebury2014 \cite{Middlebury}, SceneFlow \cite{mayer2016a_sceneflow}, and the proposed UWStereo. We train the model for 100k steps with a batch size of 16 on one RTX 4080 GPU. The crop size is set to $320\times 320$ and the learning rate is set to 0.0001. For stereo matching training, the learning rate is set to 0.0002 and the model is trained for 200k steps with a batch size of 4. We randomly crop images to $320\times 736$ and use the same data augmentation as IGEV. Following previous methods, we use 22 and 32 update iterations for training and evaluation.

\noindent\textbf{Metrics}
The widely used end-point-error (EPE) and threshold error rate ($>$3px) were used as metrics for evaluation. We set the threshold to 3px for all experiments. We also employed several image quality assessment algorithms to compare the image quality of our proposed synthetic data with the previous datasets. The image quality comparison is presented in supplementary material.

\begin{table}[ht]
\centering
\small
\centering
\caption{Ablation study for mask ratio.}
\label{tab:abla}
\begin{tabular}{l|c|cc|cc}
\toprule
\multirow{2}{*}{Model} & \multirow{2}{*}{CVE}  & \multicolumn{2}{c|}{SceneFlow}  & \multicolumn{2}{c}{UWStereo} \\
\cline{3-6}
                      &   & EPE  & $>$3px & EPE  & $>$3px   \\
\midrule
\multicolumn{6}{c}{\textit{SceneFlow} $\rightarrow$ \textit{UWStereo}} \\
\midrule
IGEV      &   &  0.52  & \textbf{2.65}  & 2.48 &  10.92  \\ 
\hline
Baseline     &  $\checkmark$  & 0.52 & 2.70 & 2.47 & 11.00   \\
\hline
0.25        &   $\checkmark$ & 0.52  & 2.72   & 2.21 & 10.18    \\
\hline
0.5      &  $\checkmark$ & \textbf{0.52} & 2.72  & \textbf{2.15} & \textbf{9.95}   \\
\hline
0.75      &  $\checkmark$ & 0.52  & 2.75  & 2.97 & 11.29   \\
\midrule
\multicolumn{6}{c}{\textit{UWStereo} $\rightarrow$ \textit{SceneFlow}} \\
\midrule
IGEV    &  &  2.14  &  \textbf{8.00}   & 0.73 & 3.45    \\ 
\hline
Baseline     &  $\checkmark$  & 2.18 & 8.38 & 0.72 & \textbf{3.29}  \\
\hline
0.25        &   $\checkmark$ & 2.21  & 8.81  & 0.73  & 3.55   \\
\hline
0.5      &  $\checkmark$ & \textbf{2.06} & 8.05  & \textbf{0.69} & 3.44  \\
\hline
0.75      &  $\checkmark$ & 2.30 &8.52  & 0.74  & 3.48    \\
\bottomrule
\end{tabular}

\end{table}

\begin{table}[]
    \centering
    \small
    \centering
    \caption{Efficiency comparison on 'Coral' scene. ELFNet and PCWNet are evaluated with $(368, 320)$ and $(640, 960)$ resolution respectively. Other methods are evaluated on $(736, 1280)$ resolution.}
    \label{tab:speed}
    \begin{tabular}{l|c|c|c}
    \toprule
    Model &  EPE & $>$3px  & Run-time \\
    \midrule
    GwcNet\cite{guo2019groupwise_gwc}    &   2.25  & 10.27  & 0.28    \\
    \hline
    ELFNet* \cite{Lou_2023_ICCV}    &  2.97  & 15.36 & 1.08    \\
    \hline
    PCWNet* \cite{shen2022pcw}   &  2.98  & 15.33 & 0.42    \\
    \hline
    ACVNet \cite{xu2022attention}    &  1.93  & 7.90 & 0.32   \\
    \hline
    RAFT-Stereo \cite{lipson2021raftstereo}    & 1.55 & 7.86 & 0.56(0.35) \\ 
    \hline
    IGEV \cite{gangwei2023iterative_igev}     &  \textbf{1.49}  & \textbf{7.53}  & 0.38(\textbf{0.22}) \\ 
    \hline
    Ours    &   1.53 & 7.91 & 0.50(0.35)  \\
    \bottomrule
    \end{tabular}
    \label{tab:runtime}
\end{table}

\subsection{Results and Analysis}\label{S5.2}
\noindent
\textbf{Benchmarking}
For benchmarking purpose, we retrain nine recent-developed models on UWStereo and present the results in Table~\ref{tab:benchmark}. We observe that the coral scene is the most difficult scene and the default scene is the easiest one. The main reason is that objects placed in the coral scene are scanned from real-world which have complex structures and appearances, increasing the difficulty for disparity regression.

The compared method could be divided into two main types named cost-filter based methods \cite{chang2018pyramid_psm, cheng2020hierarchical_lea, xu2020aanet, guo2019groupwise_gwc, xu2022attention} and iterative methods \cite{lipson2021raftstereo, gangwei2023iterative_igev, li2022practical_cres, zhao2023highfrequency_hsm}. For typical cost-filter based methods, we see that the ACVNet \cite{xu2022attention} outperforms the others by a large margin, \eg 0.85 EPE v.s. 1.49 EPE of LEAStereo when trained with all UWStereo data. The main reason is that attention based cost volume construction enhances the matching-related information.
Furthermore, we observe that iterative methods achieve superior performances than cost-filter based methods on most scenes which is mainly because iterative methods can recurrently correct the uncertain disparity prediction by iteratively retrieving local cost information. \cite{lipson2021raftstereo, gangwei2023iterative_igev, li2022practical_cres}
Among all compared methods, HSMNet \cite{zhao2023highfrequency_hsm} produces the best EPE on most scenarios probably because high-frequency features can alleviate the influence of limited image quality in underwater environments. We also observe that our proposed model outperforms IGEV in several scenes, which demonstrates the efficacy of our method in enhancing the feature representation when performing cross view feature matching.

\begin{figure*}[t]
    \centering
    \includegraphics[width=1.0\linewidth]{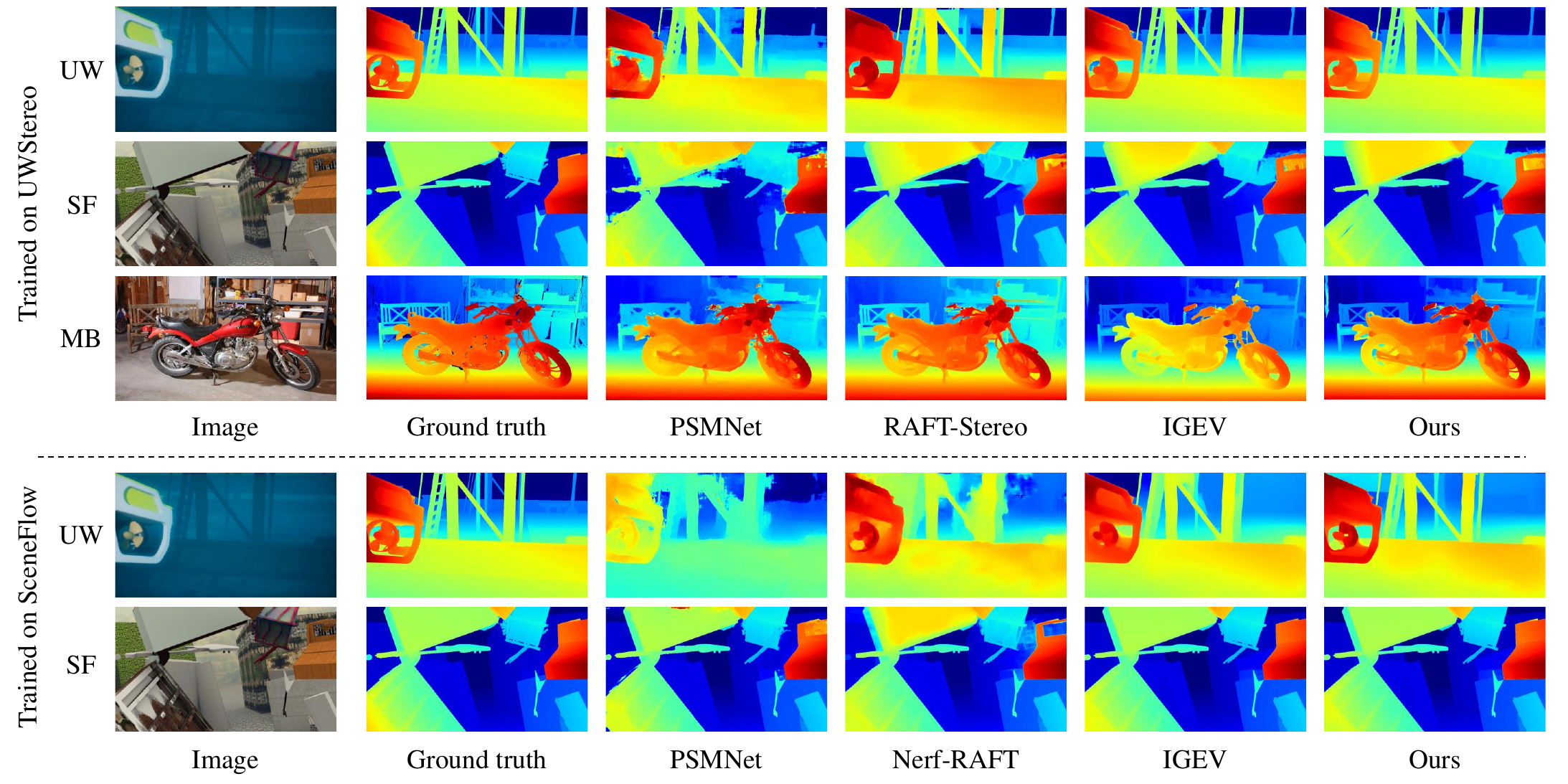}
    \caption{Visualization results. "UW", "SF", and "MB" represent UWStereo, SceneFlow, and MiddleBury2014 respectively. \textbf{Top part:} the models are trained on UWStereo and evaluated on other datasets. \textbf{Bottom part:} the models are trained on SceneFlow and evaluated on UWStereo.}
    \label{fig:visualization}
\end{figure*}

\begin{figure}[t]
    \centering
    \includegraphics[width=1.0\linewidth]{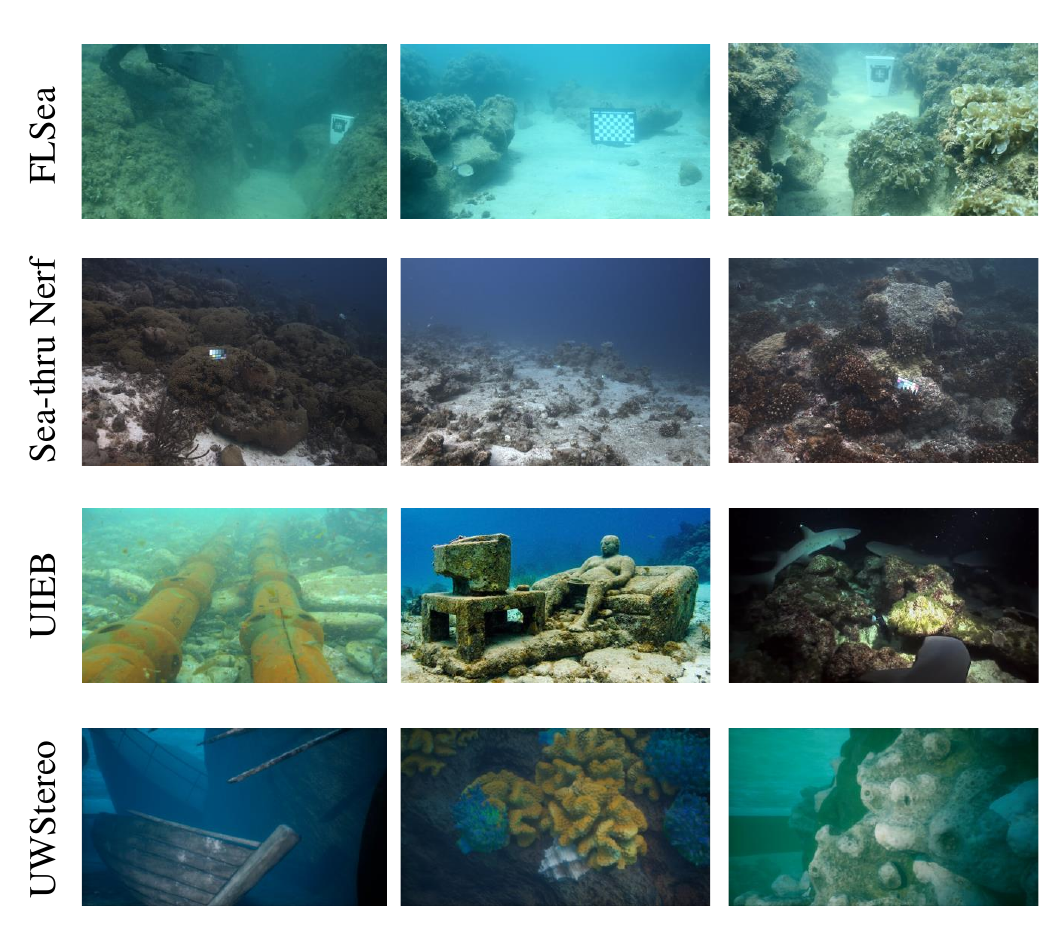}
    \caption{Comparison with underwater datasets}
    \label{fig:disc_view}
\end{figure}

\begin{figure}[t]
    \centering
    \includegraphics[width=0.9\linewidth]{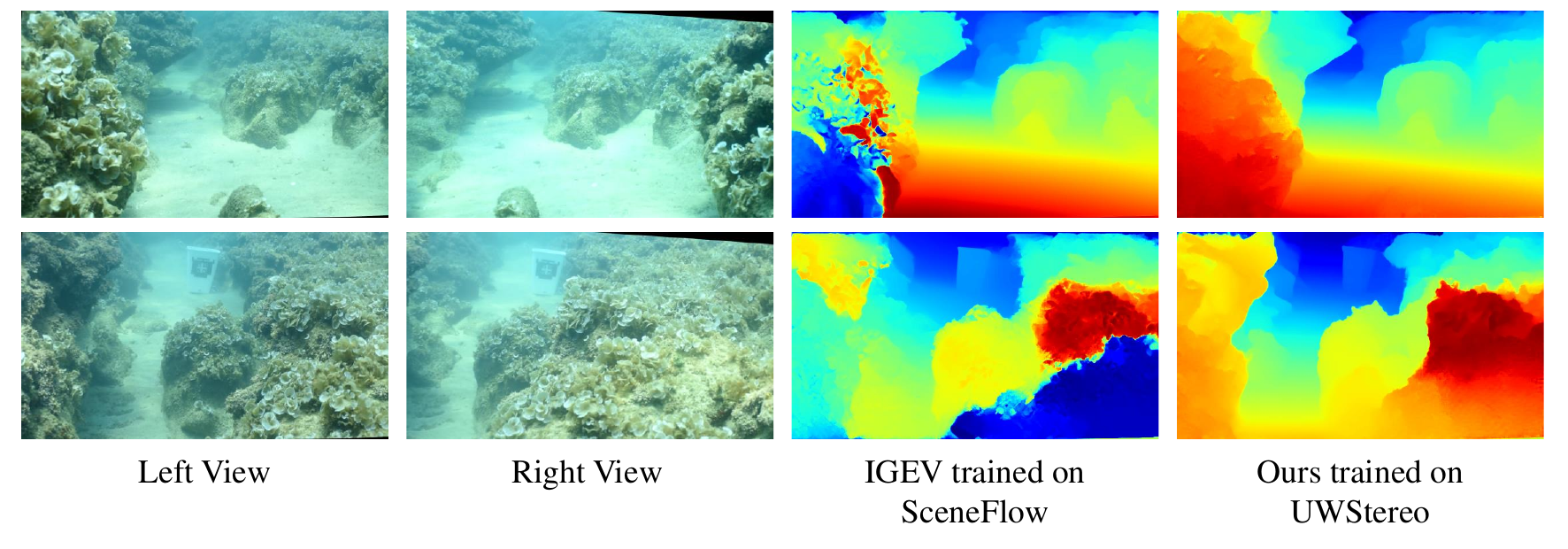}
    \caption{Real-world qualitative evaluation on FLSea-Stereo \textit{rock garden2} part.}
    \label{fig:disc_boat}
\end{figure}

\begin{figure*}
  \centering
  \begin{minipage}[t]{0.24\textwidth}
    \includegraphics[width=1.0\linewidth]{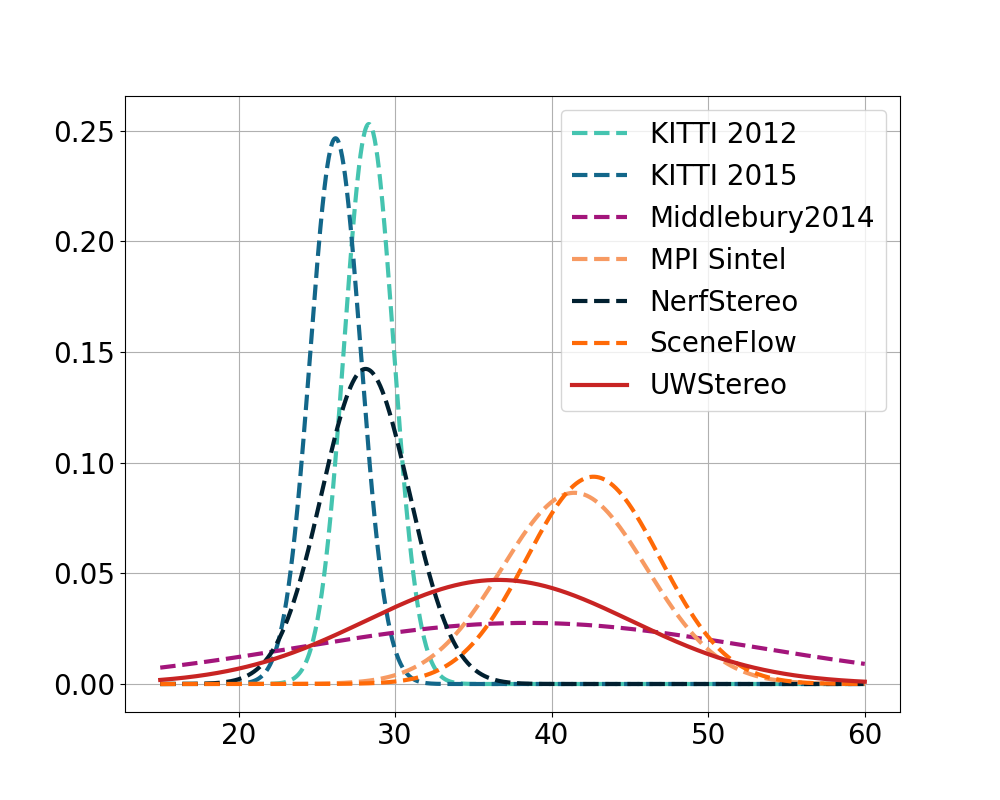}
    \caption{MUSIQ($\uparrow$).}
    \label{fig:uiqa-a}
  \end{minipage}
  \begin{minipage}[t]{0.24\textwidth}
    \includegraphics[width=1.0\linewidth]{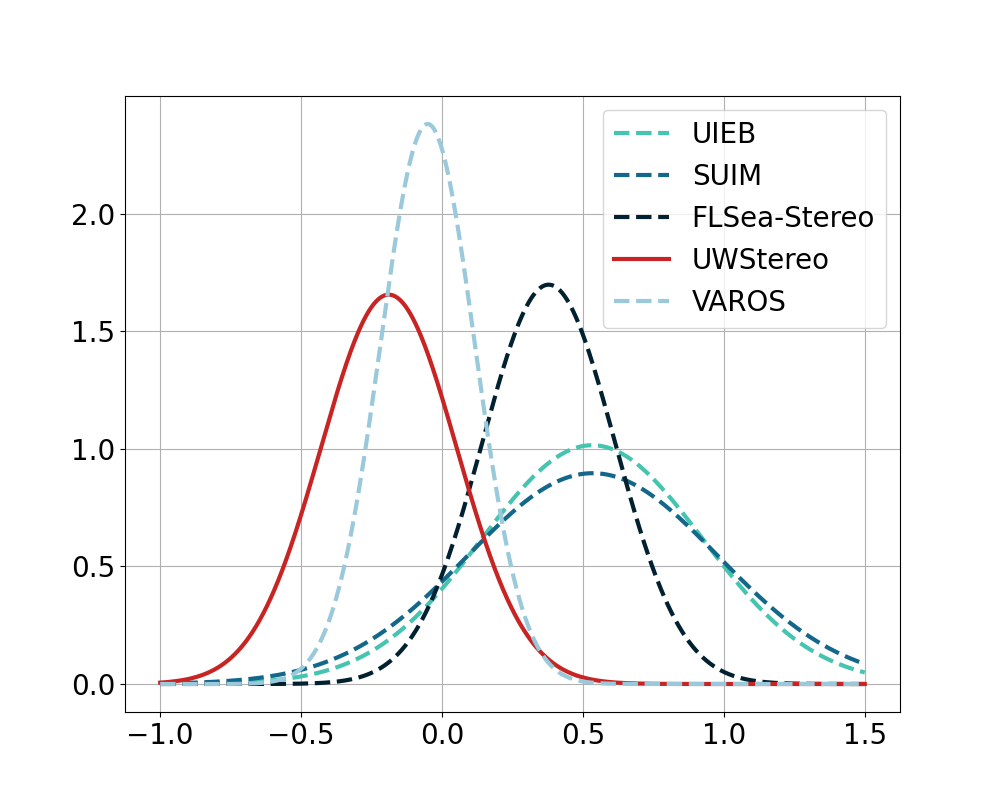}
    \caption{UIQM($\uparrow$).}
    \label{fig:uiqa-b}
  \end{minipage}
   \begin{minipage}[t]{0.24\textwidth}
    \includegraphics[width=1.0\linewidth]{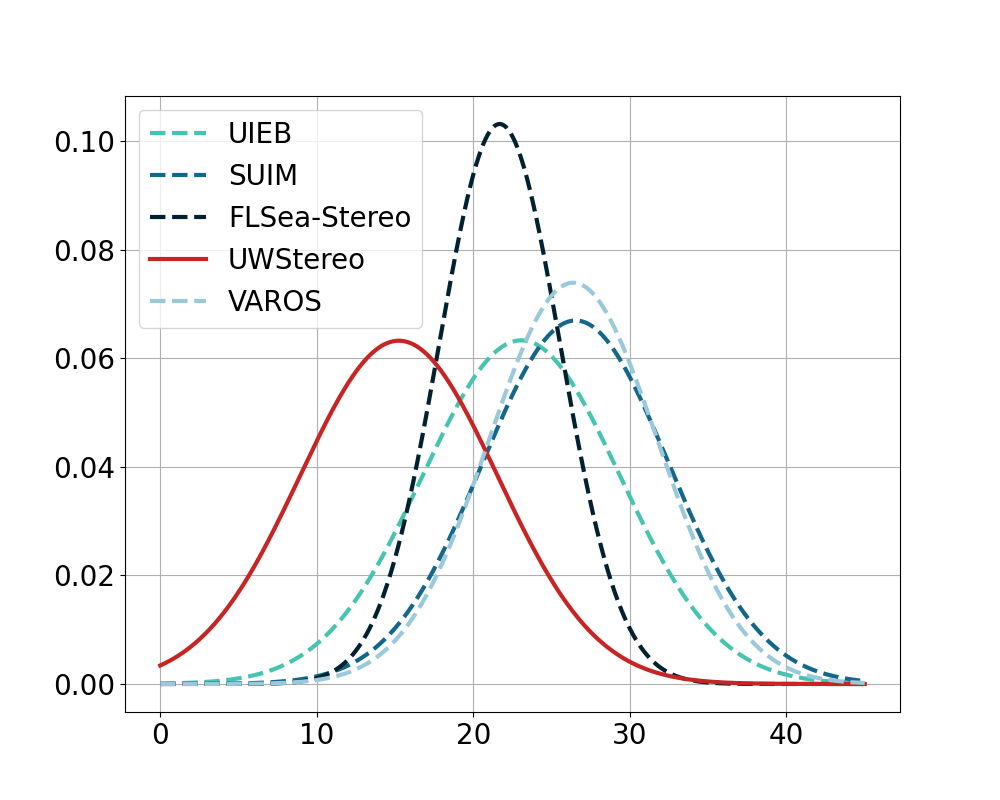}
    \caption{UCIQE($\uparrow$).}
    \label{fig:uiqa-c}
  \end{minipage}
    \begin{minipage}[t]{0.24\textwidth}
    \includegraphics[width=1.0\linewidth]{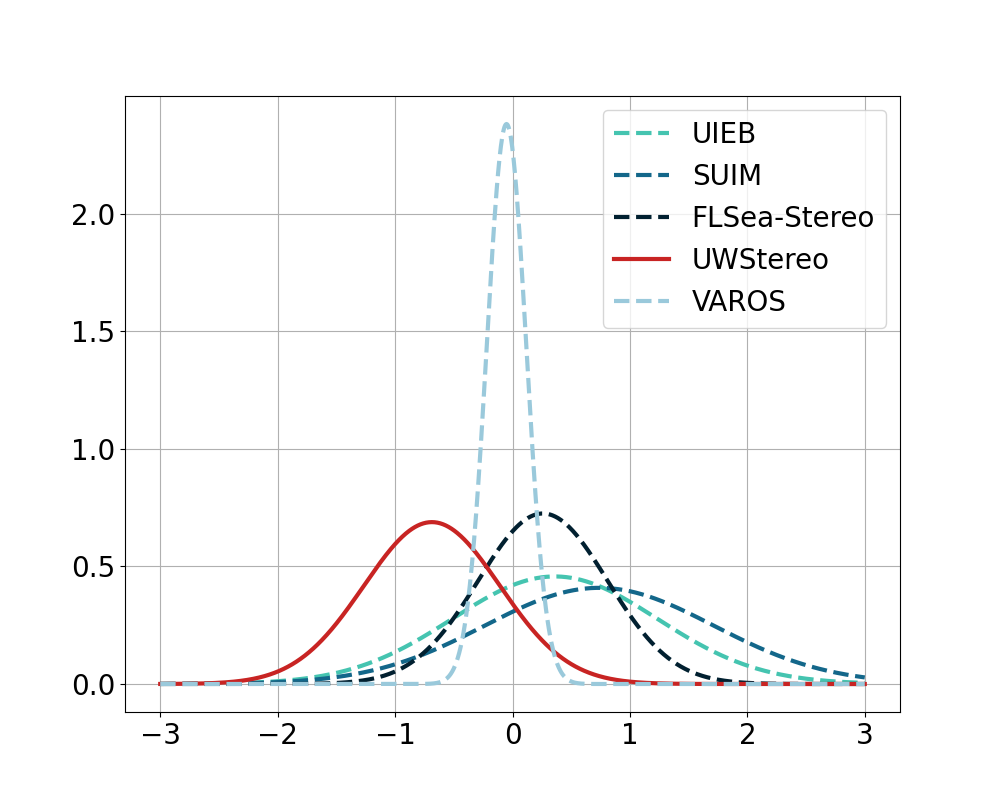}
    \caption{URanker($\uparrow$).}
    \label{fig:uiqa-d}
  \end{minipage}
\end{figure*}

\begin{figure*}[t]
    \centering
        \begin{minipage}[t]{0.32\textwidth}
    \includegraphics[width=1.0\linewidth]{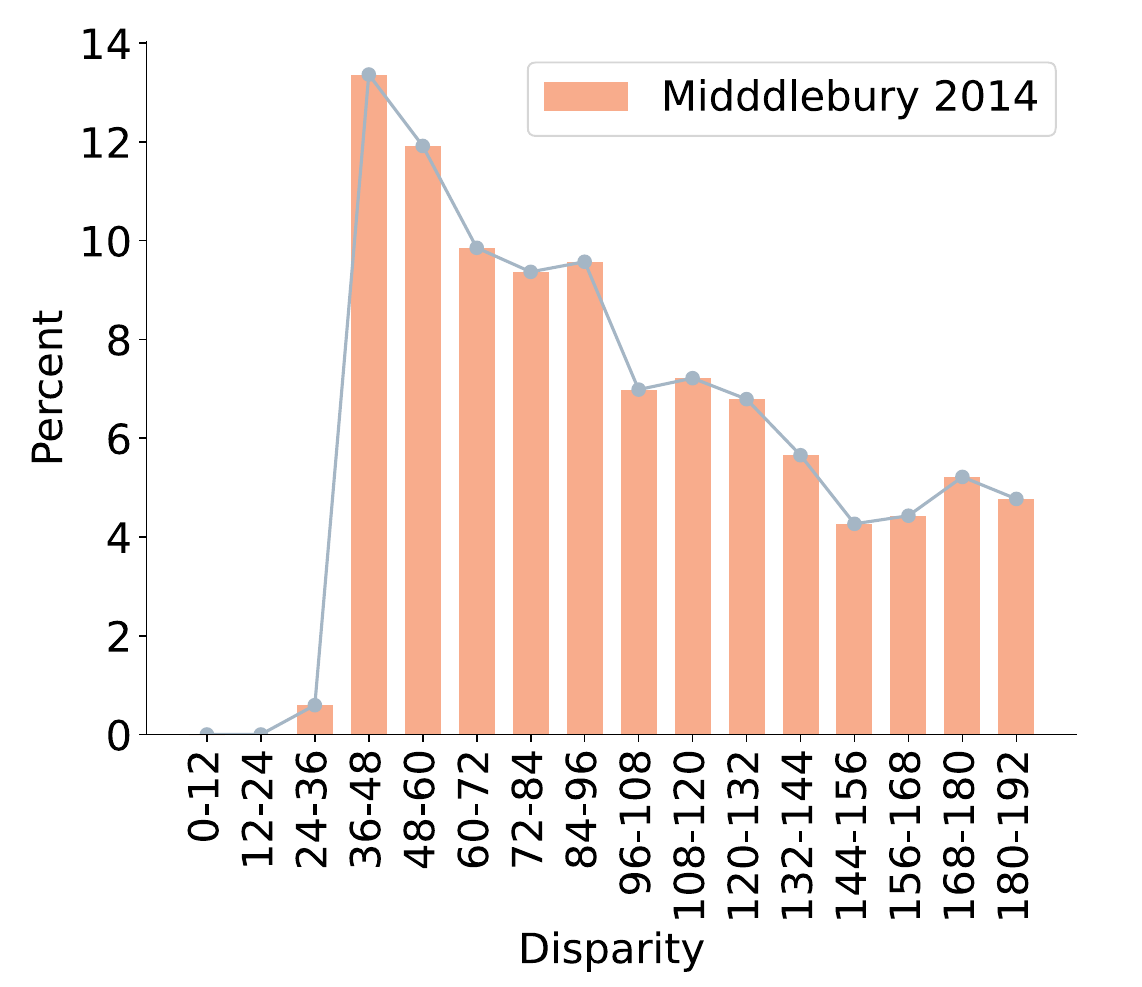}
    \caption{MiddleBury2014.}
    \label{fig:disp_mid}
    \end{minipage}
    \begin{minipage}[t]{0.32\textwidth}
    \includegraphics[width=1.0\linewidth]{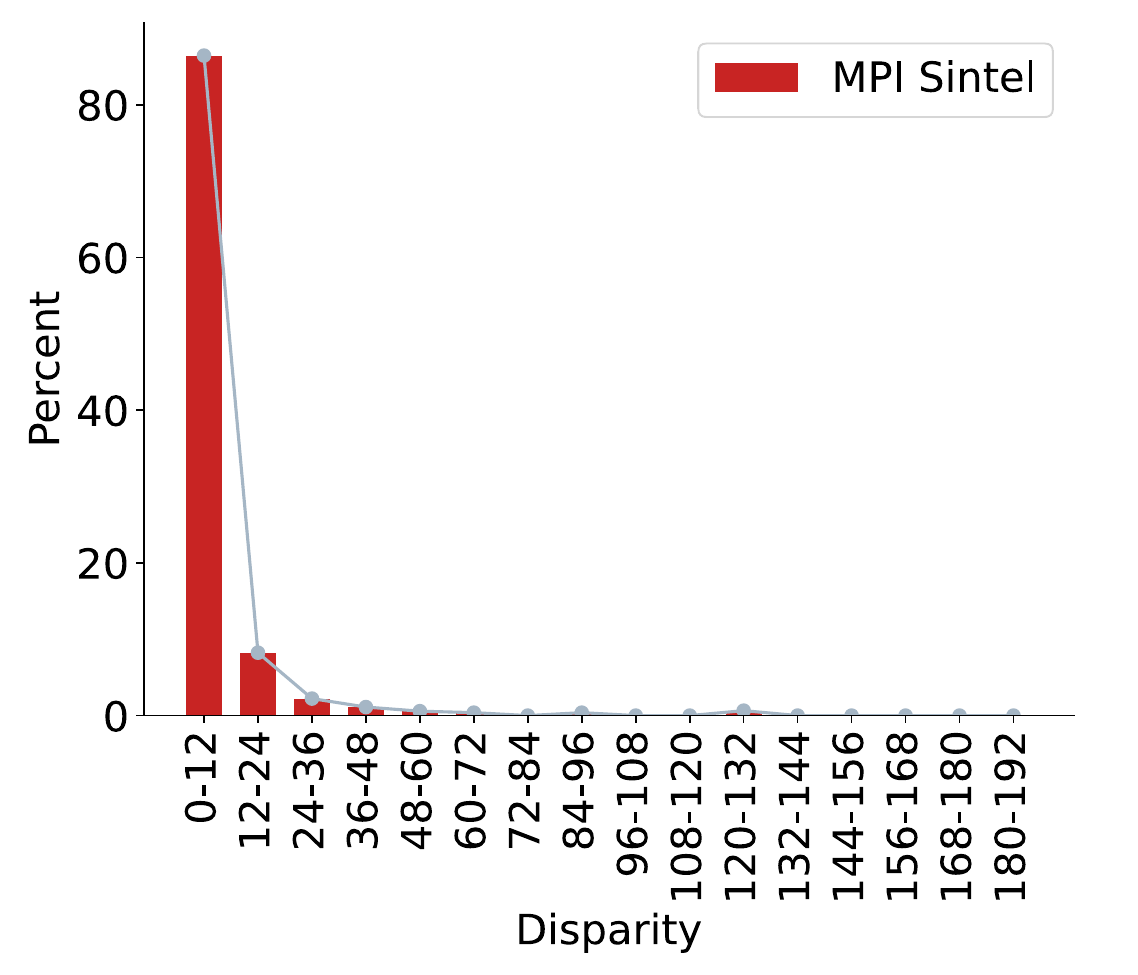}
    \caption{MPI Sintel.}
    \label{fig:disp_sintel}
    \end{minipage}
    \begin{minipage}[t]{0.32\textwidth}
    \includegraphics[width=1.0\linewidth]{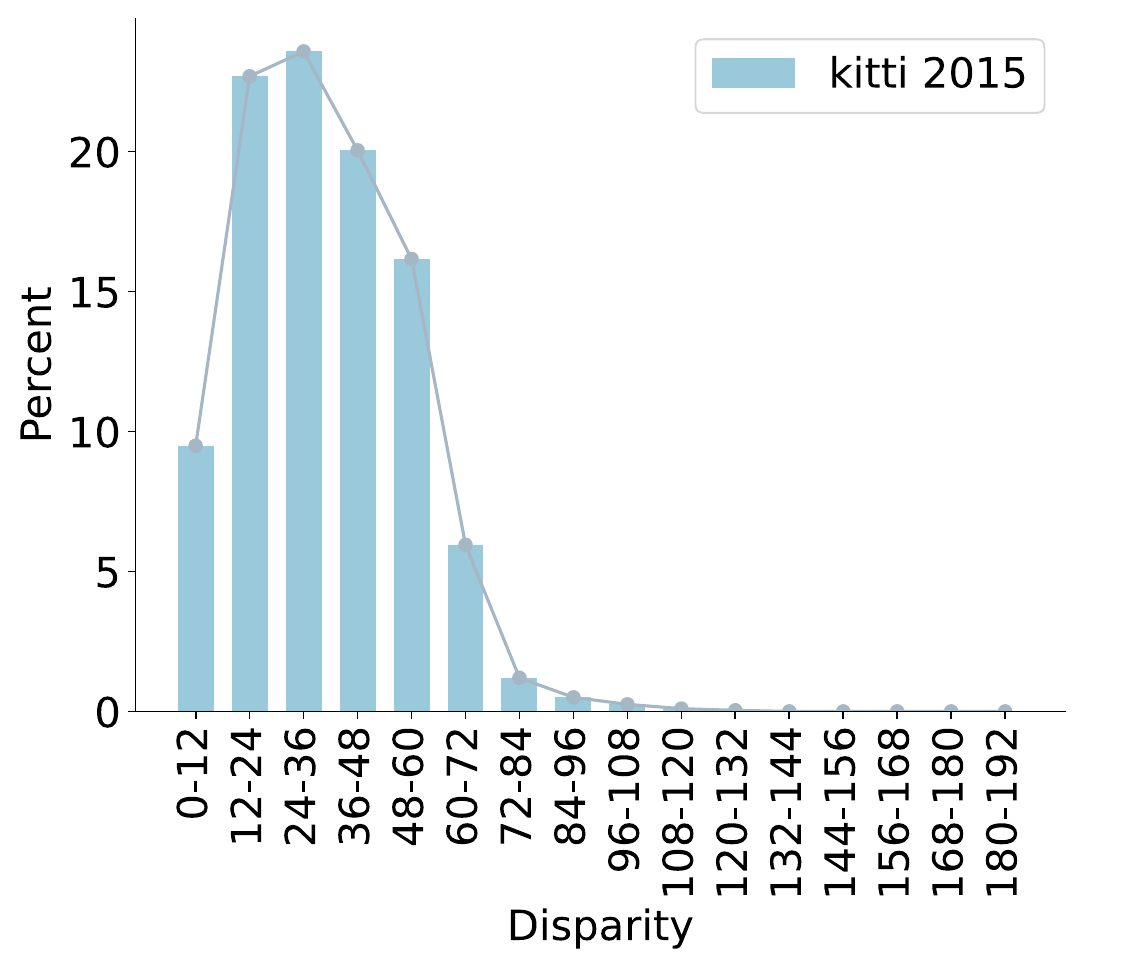}
    \caption{KITTI 2015.}
    \label{fig:disp_kitti}
    \end{minipage}

        \begin{minipage}[t]{0.32\textwidth}
    \includegraphics[width=1.0\linewidth]{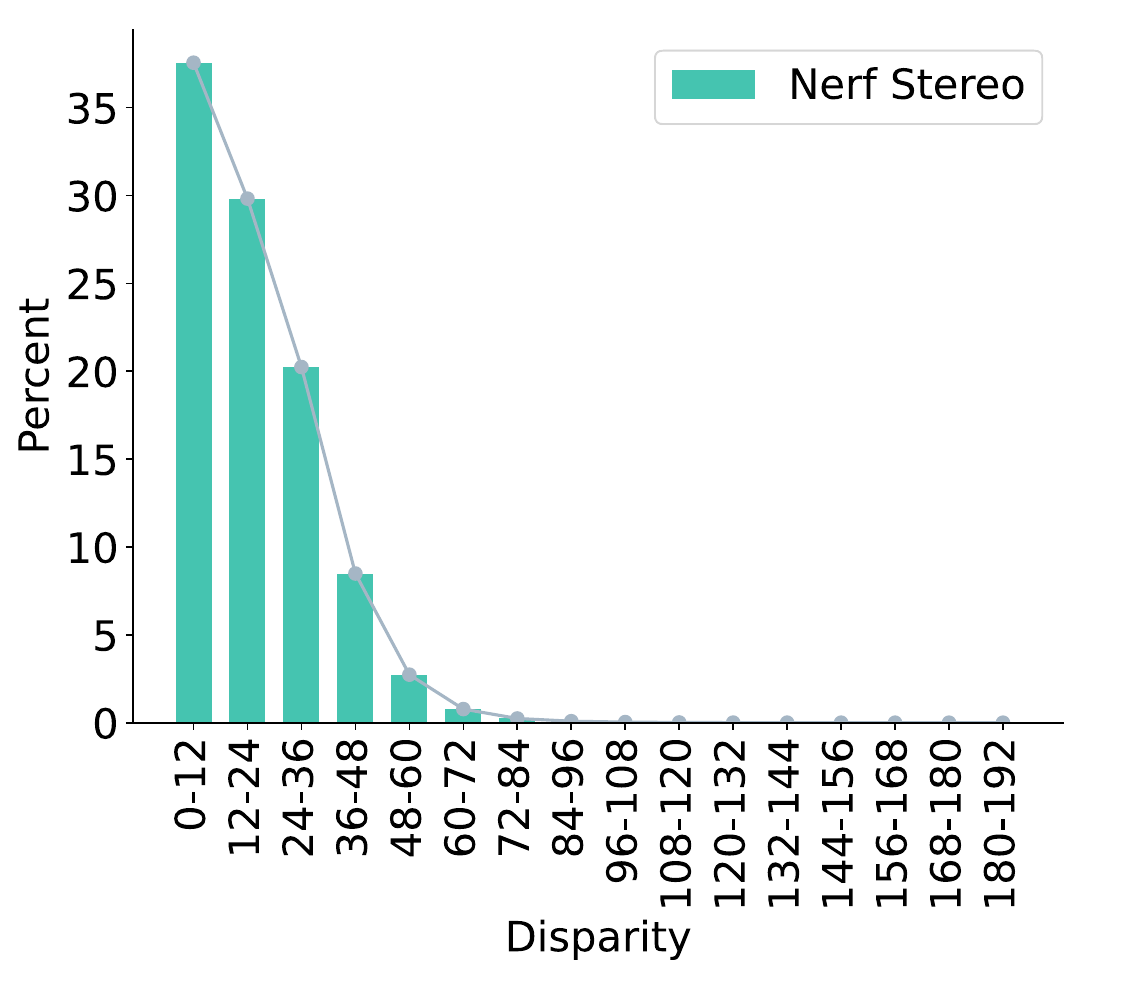}
    \caption{Nerf-Stereo.}
    \label{fig:disp_nerf}
    \end{minipage}
        \begin{minipage}[t]{0.32\textwidth}
    \includegraphics[width=1.0\linewidth]{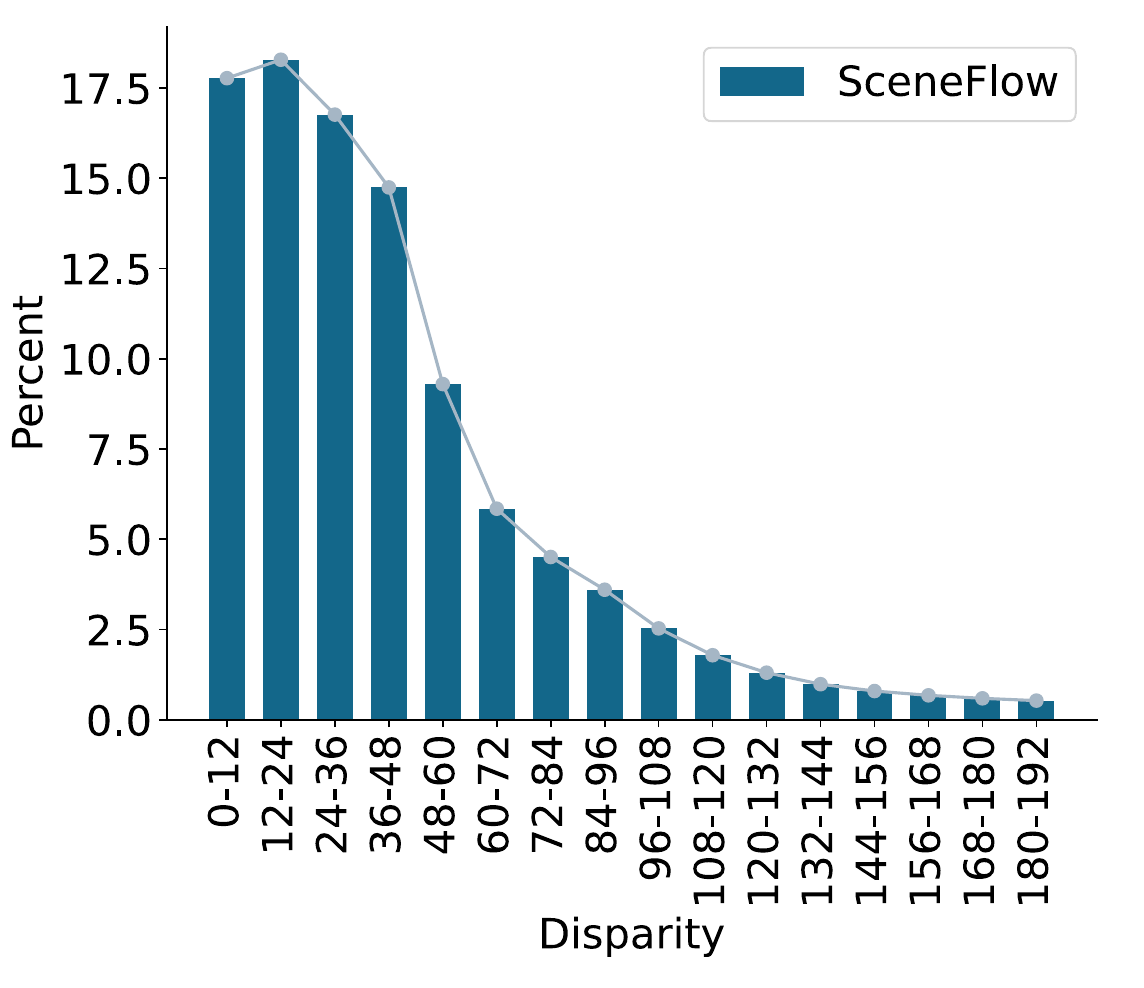}
    \caption{SceneFlow.}
    \label{fig:disp_sf}
    \end{minipage}
        \begin{minipage}[t]{0.32\textwidth}
    \includegraphics[width=1.0\linewidth]{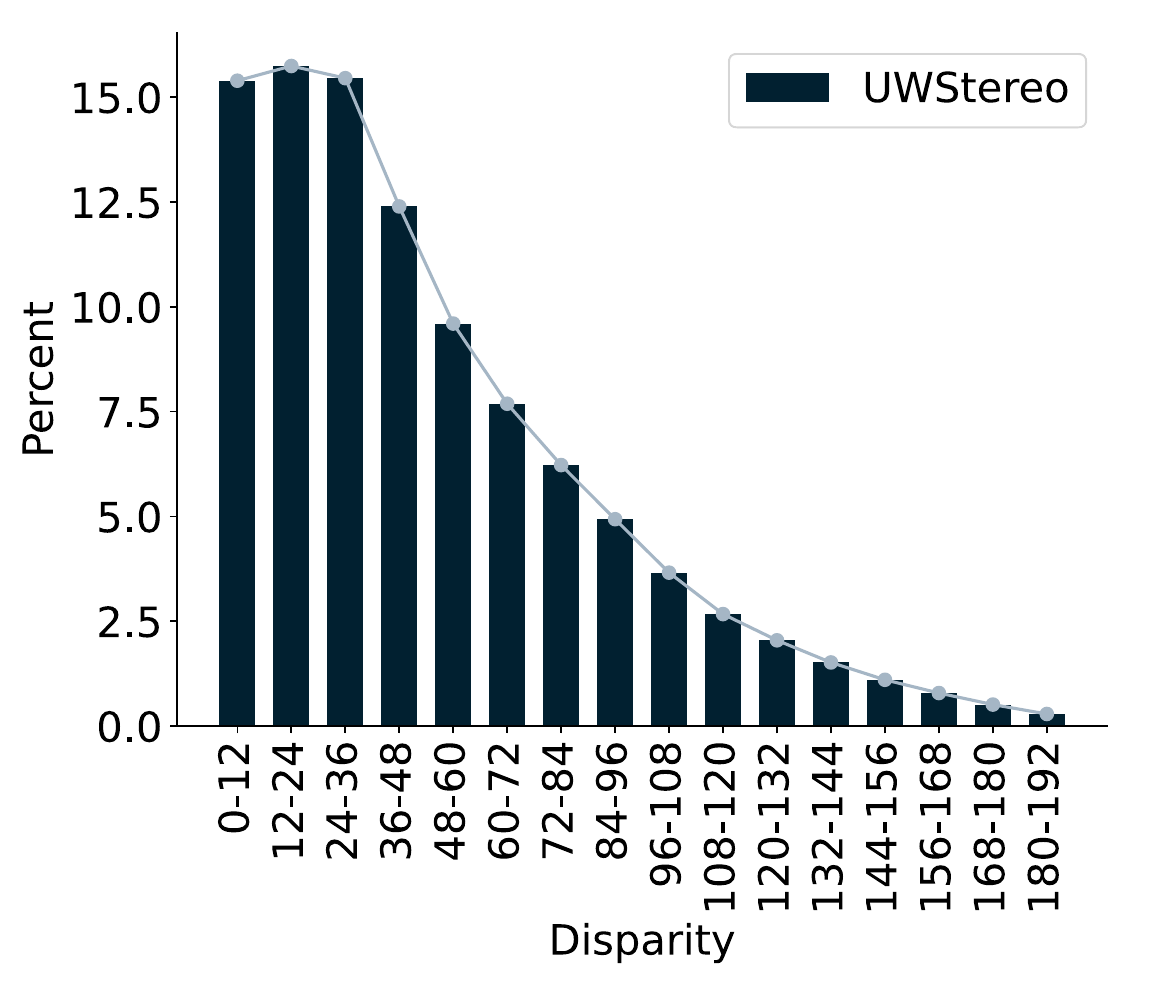}
    \caption{UWStereo.}
    \label{fig:disp_our}
    \end{minipage}
\end{figure*}

\noindent
\textbf{Generalization Benchmarking}
To study the generalization ability between terrestrial and underwater synthetic data for stereo matching models, we perform cross domain evaluation by using the weights pretrained on SceneFlow. Notably, we also take DSMNet \cite{zhang2020domaininvariant_dsm}, Graft-PSMNet \cite{liu2022graftnet}, Nerf-RAFT \cite{tosi2023nerfsupervised}, Nerf-PSMNet \cite{tosi2023nerfsupervised}, and Croco-Stereo \cite{Weinzaepfel_2023_ICCV_croco} into the comparison, because these models are designed with improved generalization ability.

As shown in Table~\ref{tab:generalizatinobenchmark}, we find that most iterative methods exhibit superior generalization performance compared to cost-filter-based methods.  However, HSMNet fails to generalize to underwater scenes, \eg 6.55 EPE on all UWStereo data. This may be attributed to the high-frequency features overfitting to a specific domain.  Another surprising observation is that the models specifically designed for generalized stereo matching may not be effective when dealing with the domain gap between terrestrial and underwater data. Concretely, Graft-PSMNet achieves 4.57 EPE and 12.82 $>$3px error rate on all UWStereo data, which is even worse than the original PSMNet. In contrast, Nerf-PSMNet demonstrates an improved EPE, decreasing from 3.54 to 3.13, likely due to the extensive Nerf-rendered images improves the generalization ability. For our method, even the model trained with limited hardware resource (e.g. we set batch size to 4 instead 8 in IGEV), it can also achieve comparable generalizaiton performances among different scenes than IGEV.

Furthermore, we observe that DSMNet and CREStereo achieve impressive generaliztion performances on several scenes. We guess the reason comes from they have been learned from similar synthetic data generated by unreal engine simulator, \eg Carla \cite{pmlr-v78-dosovitskiy17a_carla}. We also find the limited performances of Croco-stereo, which suggests that the interaction between self-supervised learning and stereo matching is still unexplored.

\noindent
\textbf{Visualization}
In Fig.~\ref{fig:visualization}, we present the visualization results of several stereo matching models by separately training the model on SceneFlow or UWStereo. In the top part, we observe that our method achieves more stable disparity estimation results when trained on UWStereo dataset than other compared methods. Furthermore, when only trained on SceneFlow (bottm part), our approach can still generalize well to underwater scene. These results demonstrate that the proposed self-supervised pretraing strategy is able to improve the cross-domain generalization ability for stereo matching networks.

\noindent
\textbf{Efficiency}
In Table~\ref{tab:speed}, we compare the efficiencies by feeding them with data with fixed $(736, 1280)$ resolution on one RTX 4080 GPU card. Due to the limitation of GPU memory, we feed ELFNet \cite{Lou_2023_ICCV} and PCWNet \cite{shen2022pcw} with $(368, 320)$ and $(640, 960)$ data. The run-time is averaged by performing 100 runs. For iterative methods, we report the metrics by using $32$ and $16$ updates. We observe that our model achieves comparable efficiency.

\noindent
\textbf{Ablation Study}
In Table~\ref{tab:abla}, we conduct an ablation study for the mask ratio by training the model on one domain and evaluating on another domain. Notably, we use small batch size for training than other methods in Table~\ref{tab:generalizatinobenchmark} due to the hardware limitation. Hence the generalization performances will degenerate to some extent. We observe our model produces best generalization performances when setting $r_1 = r_2 = 0.5$. These demonstrate that an appropriate mask ratio encourages the model to learn correspondence between left and right views. However, a higher or lower one may hinder the model to generalize to another domain.

\begin{table}[t]
\small
  \centering
\caption{Real-world quantitative evaluation on FLSea-Stereo \textit{rock garden2} part.}
\label{tab:realworld_eval}
\begin{tabular}{l|c|c}
\toprule
Model  & Training Set & Mean Error (\textit{m})   \\ 
\hline
IGEV   &  SceneFlow  & 1.10  \\
\hline
IGEV   &  UWStereo  & 1.22   \\
\hline
Our   & UWStereo  & \textbf{0.87} \\
\bottomrule
\end{tabular}
\end{table}

\subsection{Discussion}
\noindent
\textbf{The practical value of UWStereo.}
In Fig.~\ref{fig:disc_view}, we present samples from previous underwater datsets and our UWStereo. We observe that most of our samples are synthesized with close views instead of containing many distant views like FLSea \cite{randall2023flsea} or Sea-thru Nerf \cite{levy2023seathrunerf}. This implies the model trained with UWStereo may be suitable for the reconstruction of underwater objects instead of blank underwater scenes. 

For verification, we further conduct real-world quantitative and qualitative evaluations on part of FLSea-Stereo dataset, as the dataset provides estimated depth information and real-world underwater stereo images. We select the \textit{rock garden2} part for evaluation. In detail, this split totally contains $305$ image pairs. As seen in Table~\ref{tab:realworld_eval}, the results show that our method achieves a better mean error than IGEV trained with SceneFlow dataset, e.g. 0.87 v.s. 1.02. Meanwhile, when trained with the UWStereo dataset, IGEV performs worse than that trained with SceneFlow. This is mainly because simulated underwater effects hinder the learning of stereo matching models. Whereas, with our proposed pretraining strategy and CVE, the generalization ability of model for real-world underwater scene improves considerably, e.g. 1.22 v.s. 0.87. We also visualize the stereo matching results in Fig.~\ref{fig:disc_boat}, where the results suggest that our method performs more stable than IGEV trained on SceneFlow. These demonstrates that the UWStereo dataset and our method are profitable for solving real-world underwater stereo matching.

\noindent
\textbf{Detailed Disparity Distribution Comparison}
We present separate visualizations of the disparity distribution for six stereo matching datsets on Fig.\ref{fig:disp_mid}-Fig.\ref{fig:disp_our}. Notably, the MiddleBury2014 (Fig.\ref{fig:disp_mid}) dataset exhibits numerous samples of large disparity annotations, forming a distinctive pattern. In contrast, the MPI Sintel (Fig.\ref{fig:disp_sintel}) dataset displays a distribution that inversely corresponds to that of MiddleBury2014. Practical datasets, including Kitti 2015 (Fig.\ref{fig:disp_kitti}) , Nerf-Stereo (Fig.\ref{fig:disp_nerf}), and SceneFlow (Fig.\ref{fig:disp_sf}), showcase a significant concentration of disparities within the $0-72$ interval. Interestingly, our proposed UWStereo (Fig.\ref{fig:disp_our}) mirrors a similar distribution pattern to these practical datasets. This observation underscores the effectiveness of UWStereo as a valuable resource for training stereo matching models in terms of underwater environments. 

\noindent
\textbf{Image Quality Discussion}
We select four different image quality assessment methods named MUSIQ($\uparrow$) \cite{musiq}, UIQM($\uparrow$) \cite{uiqm}, UCIQE($\uparrow$) \cite{yang2015an_uciqe}, and Uranker($\uparrow$) \cite{guo2023underwater_uranker} to study the image quality. MUSIQ is designed for terrestrial data while other three methods are for underwater images. By assessing each image with a quality score, we we calculate the mean and standard deviation to fit a Gaussian distribution. Fig.~\ref{fig:uiqa-a}-Fig.~\ref{fig:uiqa-d} show all the comparisons. 

Concretely, we compare UWStereo with other five terrestrial datasets in Fig.~\ref{fig:uiqa-a},. An observation is that the qualities of synthetic data (MPI Sintel \cite{mpisintel} and SceneFlow \cite{mayer2016a_sceneflow}) are better than those of real-world data (KITTI \cite{kitti2012, kitti2015}, MiddleBury2014 \cite{Middlebury}). While, the qualities of Nerf-rendered images \cite{tosi2023nerfsupervised} are similar to those captured in real-world scenes. As for UWStereo, we see a medium image quality level which lies in the location that is higher than real-world data but lower than previous synthetic data. This demonstrates that our UWStereo is closer to real-world data than previous synthetic datasets. 

Fig.~\ref{fig:uiqa-b}, Fig.~\ref{fig:uiqa-c}, and Fig.~\ref{fig:uiqa-d} show the comparisons between our UWStereo and other four underwater datasets (\eg UIEB \cite{uieb}, VAROS \cite{zwilgmeyer2021the_varos}, HIMB \cite{himbdataset}, and Flsea-Stereo \cite{randall2023flsea}). Despite VAROS produces similar performances in UCIQE and URanker, the discrepancy between synthetic data and real-world underwater data is still exists because the rendering engine cannot perfectly simulate real-world visual effects. But since there is no effective way to acquire accurate depth information for real-world underwater environments, our UWStereo can be considered as a substitute for a real-world dataset to facilitate the future researches for underwater stereo matching task.

\section{Conclusion}

In this paper, we present UWStereo, a comprehensive synthetic dataset designed specifically for underwater stereo matching. This dataset features a wide variety of objects and environmental variations, and it includes 29,568 stereo image pairs meticulously annotated with dense disparity maps for the left view. To assess the robustness of current approaches, we benchmark nine state-of-the-art stereo matching methods against UWStereo and analyze their ability to generalize to underwater scenes. The results reveal that existing models face significant challenges when applied to underwater environments, highlighting a critical gap in the field. To address these challenges, we draw inspiration from masked image learning and propose a novel cross-view enhancement module. This module introduces a new training strategy that focuses on reconstructing cross-domain masked images prior to stereo matching training, thereby enhancing the models' generalization capabilities. We also explore the practical implications of our proposed dataset, discussing how it can serve as a valuable resource for the development of more robust underwater stereo matching algorithms.

In conclusion, UWStereo represents a significant advancement in the field of underwater stereo matching, offering extensive support for both current research and future innovation. By providing a large-scale, well-annotated dataset tailored to the unique challenges of underwater environments, UWStereo paves the way for the development of more effective and generalized stereo matching solutions in this specialized domain.

\bibliographystyle{bibli/IEEEtran}
\bibliography{bibli/uwstereo}

\section{Biography Section}
 
\begin{IEEEbiography}[{\includegraphics[width=1in,height=1.25in,clip,keepaspectratio]{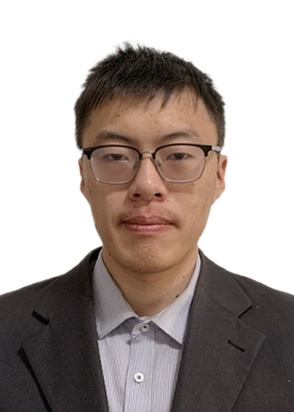}}]{Qingxuan Lv}
was born in Shanxi, China, in 1996. He received his bachelor's degree in Computer Science and Technology from the Shanxi University of Finance and Economics in 2018. He received his master's degree in Computer Science and Technology from the Ocean University of China (OUC) in 2021. He is currently a candidate of a doctor's degree at the ocean group of VisionLab OUC. His research interests include computer vision and machine learning. Specifically, he is interested in universal domain adaptation and semantic segmentation.
\end{IEEEbiography}

\begin{IEEEbiography}[{\includegraphics[width=1in,height=1.25in,clip,keepaspectratio]{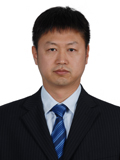}}]{Junyu Dong} received the B.Sc. and M.Sc. degrees in applied mathematics from the Department of Applied Mathematics, Ocean University of China, Qingdao, China, in 1993 and 1999, respectively, and the Ph.D. degree in image processing from the Department of Computer Science, Heriot-Watt University, Edinburgh, U.K., in November 2003. He is currently a Professor and the Head of the Department of Computer Science and Technology. His research interests include machine learning, big data, computer vision, and underwater image processing.
\end{IEEEbiography}

\begin{IEEEbiography}[{\includegraphics[width=1in,height=1.25in,clip,keepaspectratio]{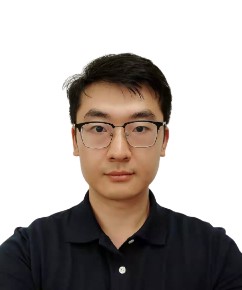}}]{Yuezun Li} (Member, IEEE) received the B.S. degree in software engineering from Shandong University in 2012, the M.S. degree in computer science in2015, and the Ph.D. degree in computer science from University at Albany–SUNY, in 2020. He was a Senior Research Scientist with the Department of Computer Science and Engineering, University at Buffalo–SUNY. He is currently a Lecturer with the Center on Artifcial Intelligence, Ocean University of China. His research interests include computer vision and multimedia forensics. His work has been published in peer reviewed conferences and journals, including ICCV, CVPR, TIFS, TCSVT, etc. 
\end{IEEEbiography}

\begin{IEEEbiography}[{\includegraphics[width=0.9in,height=1.1in,clip,keepaspectratio]{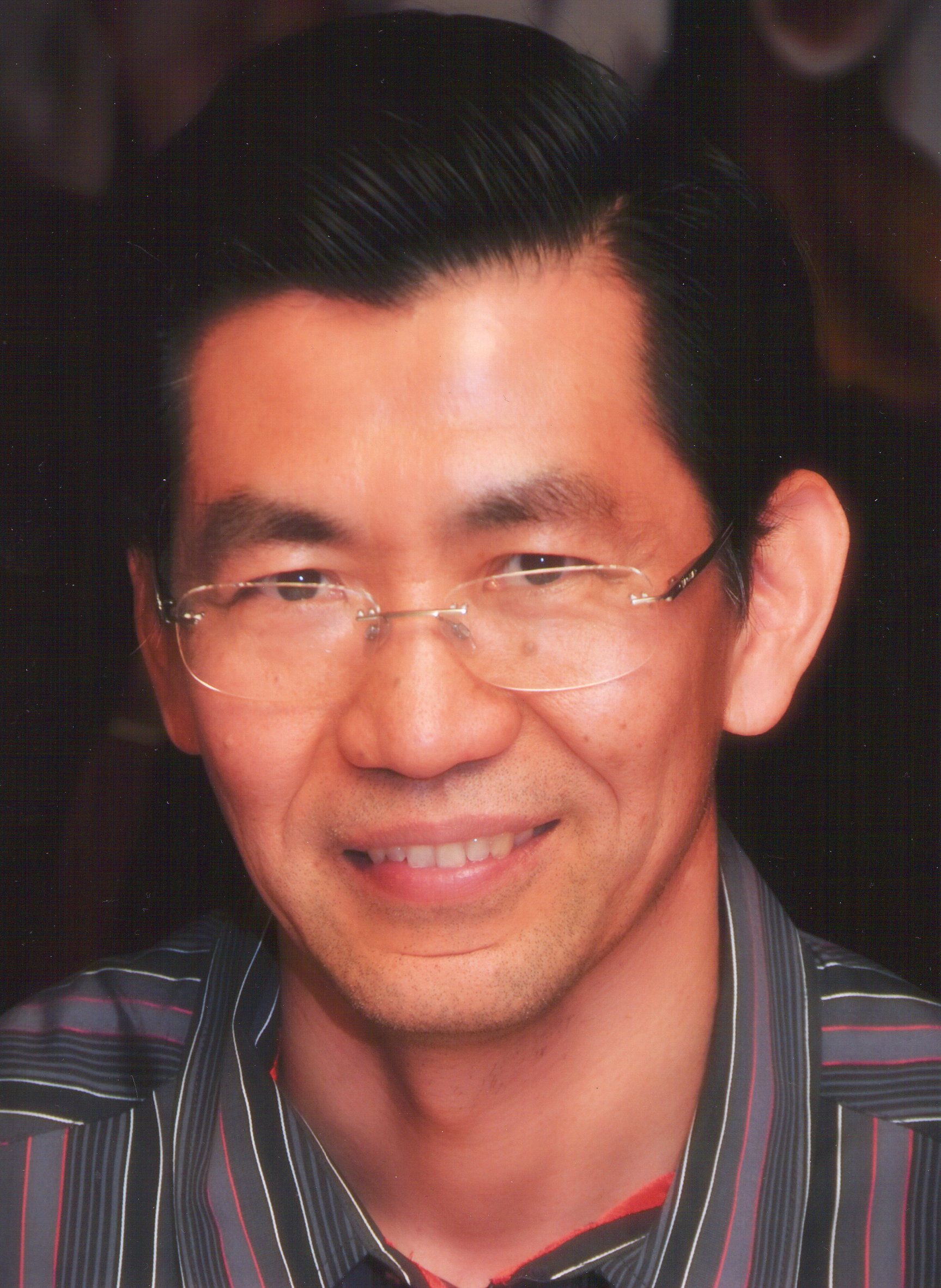}}]{Sheng Chen} (IEEE Life Fellow) received the B.Eng. degree in control engineering from the East China Petroleum Institute, Dongying, China, in 1982, the Ph.D. degree in control engineering from City University, London, in 1986, and the higher doctoral (D.Sc.) degree from the University  of Southampton, Southampton, U.K., in 2005. From 1986 to 1999, he held research and academic appointments with the University of Sheffield, the University of Edinburgh, and the University of Portsmouth, U.K. Since 1999, he has been with the School of Electronics and Computer Science, University of Southampton, where he is a Professor of Intelligent Systems and Signal Processing. His research interests include adaptive signal processing, wireless communications, modeling and identification of nonlinear systems, neural network and machine learning, intelligent control system design, evolutionary computation methods, and optimization. Professor Chen has authored over 700 research papers. He have 20,000+ Web of Science citations with h-index 61, and 39,000+ Google Scholar citations with h-index 83. Dr Chen was elected to a fellow of the United Kingdom Royal Academy of Engineering in 2014. He is a fellow of Asia-Pacific Artificial Intelligence Association (FAAIA), a fellow of IET, and an original ISI Highly Cited Researcher in engineering (March 2004).

\end{IEEEbiography}

\begin{IEEEbiography}[{\includegraphics[width=1in,height=1.25in,clip,keepaspectratio]{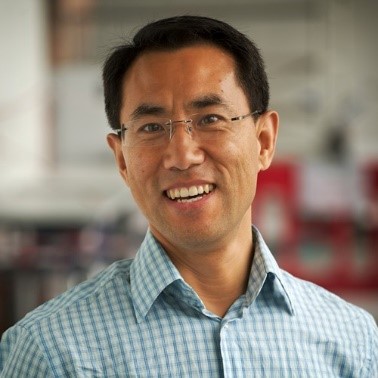}}]{Hui Yu} received the PhD degree from Brunel University London in 2009. His research interests include visual and cognitive computing, social vision, social robot and machine learning. His research particularly focuses on 3D/4D facial expression reconstruction and perception, image and video analysis for human-machine and social interaction as well as intelligent vehicle applications. He leads the Visual Computing and Social Robot Group (VCSR) in cSCAN at the University of Glasgow. He has been awarded the Industrial Fellowship project by the Royal Academy of Engineering. He also serves as an Associate Editor for Neurocomputing, IEEE Transactions on Human-Machine Systems, IEEE Transactions on Intelligent Vehicles, and IEEE Transactions on Computational Social Systems journal. More publication information can be found on Google Scholar.
\end{IEEEbiography}

\begin{IEEEbiography}[{\includegraphics[width=1in,height=1.25in,clip,keepaspectratio]{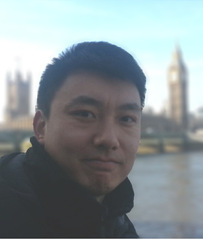}}]{Shu Zhang} 
Shu Zhang is currently an Associate Professor and Postgraduate supervisor at Ocean University of China, Qingdao, China. He received his PhD in Computer Application Technologies from Ocean University of China. He was previously a research associate at the University of Portsmouth, Portsmouth, UK. His main research interests include computer vision, feature analysis, 3D reconstruction, video processing, underwater image analysis, and deep learning among others. 
\end{IEEEbiography}

\vspace{11pt}

\begin{IEEEbiography}[{\includegraphics[width=1in,height=1.25in,clip,keepaspectratio]{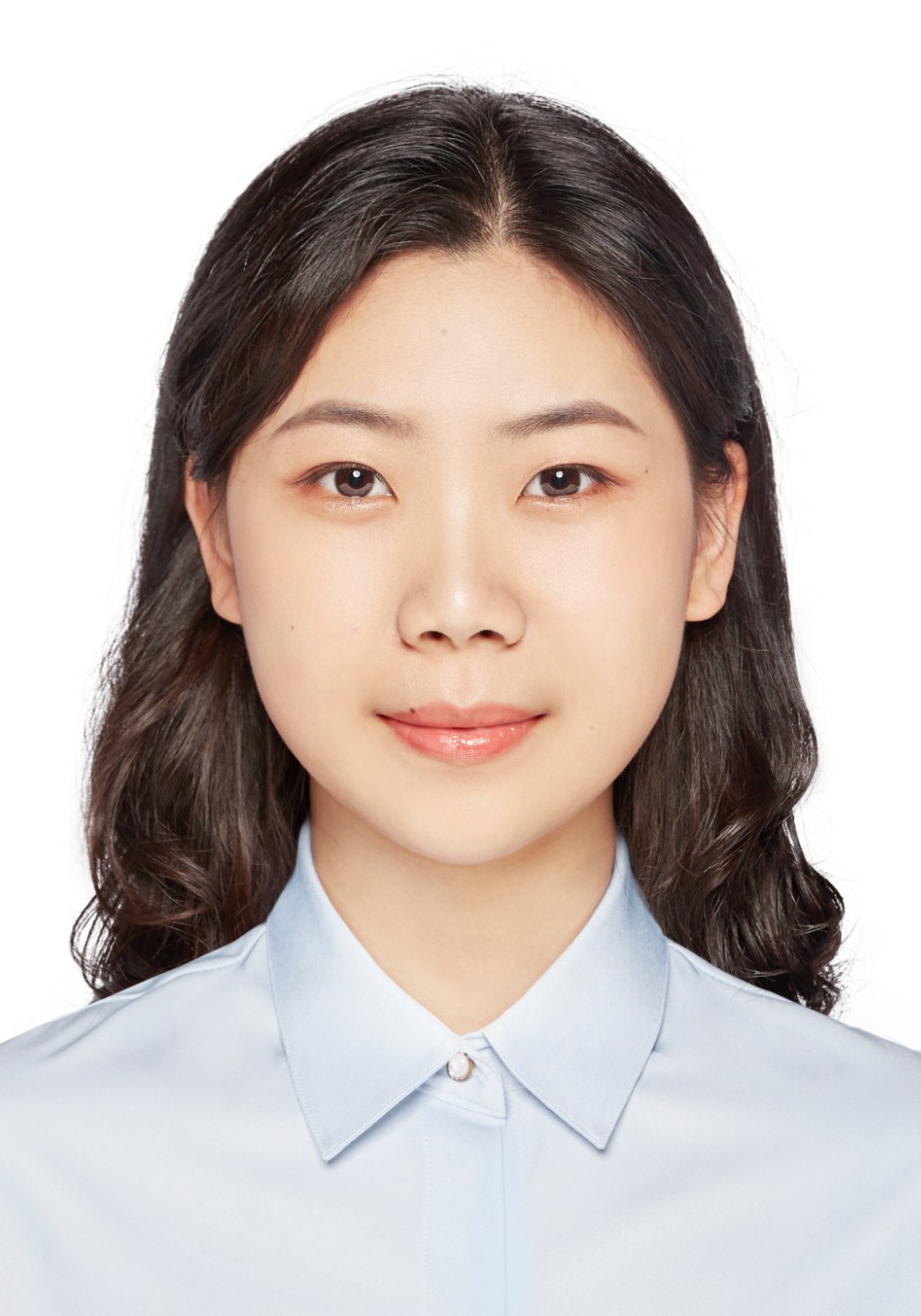}}]{Wenhan Wang} 
Wenhan Wang is a postgraduate student majoring in computer technology at Ocean University, China, Qingdao, China. Her main research interests include computer vision, SLAM, underwater image analysis and among others.
\end{IEEEbiography}

\end{document}